\documentclass{article}

\usepackage{microtype}
\usepackage{graphicx}
\usepackage{subcaption}
\usepackage{booktabs} % for professional tables
\usepackage{colortbl}
\usepackage{xcolor}         % colors
\usepackage{bm}
\definecolor{lightblue}{HTML}{F0F8FF}
\definecolor{softblue}{rgb}{0.21,0.49,0.74}
\definecolor{badred}{RGB}{200, 0, 0}    % 红色：性能下降/Token增加
\definecolor{goodgreen}{RGB}{0, 150, 0} % 绿色：性能提升/Token减少

\usepackage{hyperref}

\usepackage{multirow}
\usepackage{makecell} % 核心宏包

\usepackage[accepted]{icml2026}

\usepackage{amsmath}
\usepackage{amssymb}
\usepackage{mathtools}
\usepackage{amsthm}

\usepackage[capitalize,noabbrev]{cleveref}

\theoremstyle{plain}

\theoremstyle{definition}

\theoremstyle{remark}

\usepackage[textsize=tiny]{todonotes}

\icmltitlerunning{VisionPulse: Dynamic Visual Sparsity for Efficient Multimodal Reasoning}

\begin{document}

\twocolumn[
  % \icmltitle{Submission and Formatting Instructions for \\
  %   International Conference on Machine Learning (ICML 2026)}

  \icmltitle{VisionPulse: Dynamic Visual Sparsity for Efficient Multimodal Reasoning}



  \icmlsetsymbol{equal}{*}

  \begin{icmlauthorlist}
    \icmlauthor{Hengbo Xu}{ruc}
    \icmlauthor{Shengjie Jin}{ruc}
    \icmlauthor{Yanbiao Ma}{ruc}
    \icmlauthor{Zhiwu Lu}{ruc}
  \end{icmlauthorlist}
  
  \icmlaffiliation{ruc}{Gaoling School of Artificial Intelligence, Renmin University of China, Beijing, China}

  \icmlcorrespondingauthor{Zhiwu Lu}{luzhiwu@ruc.edu.cn}
    % \icmlcorrespondingauthor{Yanbiao Ma}{ybma1998@ruc.edu.cn}
  % \icmlcorrespondingauthor{Hengbo Xu}{hengboxu@ruc.edu.cn}

  \icmlkeywords{Computer Vision, Large Multimodal Models, Multimodal Reasoning}

  \vskip 0.3in
]



\printAffiliationsAndNotice{}  
% Zhiwu Lu is the corresponding author.
% no special notice (required even if empty)

\begin{abstract}

With the rapid advancement of large multimodal models (LMMs), inference-time overhead has become a key bottleneck for real-world deployment. 
Existing methods typically prune visual tokens at prefill, assuming the required visual evidence remains static during reasoning. 
However, we empirically show that visual evidence is strongly step-dependent: only a sparse subset of visual tokens is critical at each decoding step, and the critical set evolves across reasoning.
Furthermore, we identify a coupled bottleneck where redundant visual context can steer the model toward query-irrelevant regions, lengthening the reasoning trace.
Guided by these insights, we propose \textbf{VisionPulse}, a step-wise visual token pruning framework during reasoning. 
VisionPulse computes a lightweight visual attention mass to estimate the step-wise retention budget by exploiting its strong positive correlation with LMMs' effective visual token usage and retain only the most critical tokens under this budget. 
By enforcing visual sparsity during reasoning, VisionPulse filters redundant visual context while preserving relevant visual evidence, shortening reasoning traces naturally.
Extensive experiments show that VisionPulse only retains 5\% of visual tokens per step with reasoning traces shortened by 11.2\%, while keeping accuracy almost unchanged.

\end{abstract}

\begin{figure*}[t]  
    \centering
    \centerline{\includegraphics[width=0.96\linewidth]{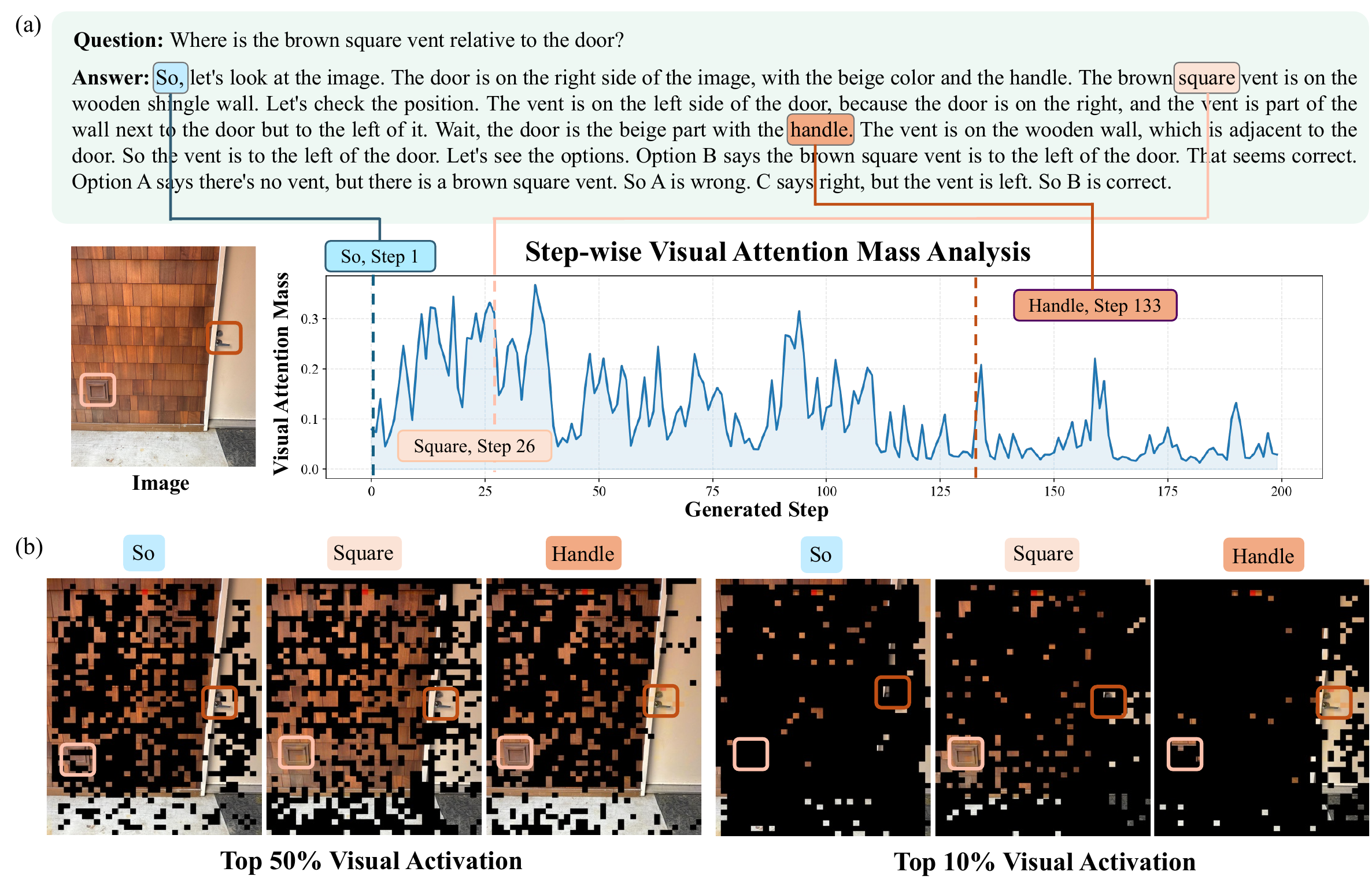}}
    \vspace{-0.35cm}
    \caption{\textbf{Dynamic visual activations during multimodal reasoning.}
     \textbf{(a) Step-wise Visual attention mass over decoding.} 
     We measure the step-wise visual attention mass (total attention allocated to visual tokens) during reasoning. 
     Visual evidence is strongly step-dependent: it remains negligible in text-dominated steps, but increases when the reasoning involves referenced entities (e.g., \emph{square}, \emph{handle}).
    \textbf{(b) Visual attention heatmap at representative steps.} Top-50\% and top-10\% visual-attention activation masks show that attention concentrates on regions corresponding to referenced entities at high-mass steps, but remains diffuse at low-mass steps. This indicates that the amount and the subset of critical visual tokens vary during reasoning.
    }
    \label{figure_observation}  
   \vskip -0.2in
\end{figure*}

\vspace{-0.02in}
\section{Introduction}

Large multimodal models (LMMs)~\cite{openaio1,qwen3vl,internvl3} have recently achieved substantial progress in multimodal reasoning, enabling vision-centric, multi-step chain-of-thought (CoT) reasoning~\cite{wang2025vlrethinker,yang2025r1onevision,perceptionsurvey, multimodalsurvey, sun2025cotmr, dai2026harder} across tasks such as scientific problem solving, chart understanding, and embodied perception. 
However, as reasoning capabilities continue to improve, inference-time overhead has emerged as a primary bottleneck for real-world deployment~\cite{vllm,lin2024awq}.

Under existing multimodal decoding paradigms, most LMMs~\cite{li2024llava,qwen3vl,internvl3,yao2024minicpm} treat image and text tokens uniformly within a shared self-attention mechanism, where each decoding step computes interactions over the entire multimodal context.
On the visual side, the strong performance of LMMs depends on fine-grained visual representations.
As visual understanding demands grow, images and videos are represented more densely, resulting in substantially more visual tokens and higher computational cost.
On the reasoning side, as LMMs become stronger multimodal reasoners, they are increasingly able to tackle complex, out-of-distribution tasks.
Meanwhile, when faced with hard queries, LMMs typically generate longer multi-step reasoning traces, which expand the language context during decoding.
Consequently, each decoding step must attend to both an expanding language context and a large visual token budget, leading to a sharp increase in inference overhead.

A natural question thus arises in multimodal reasoning: are all visual tokens and generated reasoning tokens equally necessary throughout the entire reasoning process?
Prior studies~\cite{wang2021not,liang2022not,yu2022unified} have noted that the information content in images is often sparse.
Moreover, for a given query, only a limited subset of visual information is actually relevant, motivating a range of visual compression methods for efficient multimodal inference~\cite{visionzip,fastv,sparsevlm,mllmsknow,vflowopt}.
These methods typically perform compression at the prefill stage by selecting a fixed subset of visual tokens for subsequent decoding.
While effective for direct-answer generation, this design implicitly assumes that the selected visual tokens remain relevant and unchanged throughout the reasoning process.

To examine this assumption for multimodal reasoning, we conduct an empirical analysis of multimodal reasoning models and focus on how visual tokens are utilized.
Specifically, we quantify the aggregate attention allocated to visual tokens at each decoding step (i.e., visual attention mass) and analyze the corresponding attention heatmaps for representative decoding steps.
As shown in~\cref{figure_observation}, the need for visual evidence is activated on demand as the reasoning state evolves. 
It is negligible in text-dominated steps but increases substantially when the model must ground specific visual entities.
Moreover, LMM's visual attention shifts across regions over time, indicating that the critical visual tokens vary during reasoning.
Therefore, no single fixed subset can remain optimal throughout reasoning.

These observations thus expose a fundamental mismatch in existing static pruning for efficient multimodal reasoning.
Most prior methods fix a single visual subset once at the start of decoding, when the model generates the first token and allocates little attention mass to visual tokens, as shown in~\cref{figure_observation}.
As a result, a fixed subset is inevitably misaligned with later demands: it may discard visual tokens that become crucial at subsequent steps that require visual evidence, while still preserving redundant context during text-dominated reasoning.
This mismatch becomes particularly pronounced under aggressive retention, where early pruning errors cannot be corrected and may propagate through the entire reasoning process.

Motivated by these observations, we propose \textbf{VisionPulse}, a \textbf{training-free} dynamic visual pruning framework for multimodal reasoning.
Unlike prior methods~\cite{fastv,sparsevlm,pyramiddrop} that prune visual tokens once at prefill, VisionPulse performs step-wise visual token selection during reasoning, enabling the model to select critical visual tokens as the reasoning state evolves.
Beyond preserving accuracy, we uncover a coupled bottleneck: retaining redundant visual context at every step steers LMMs toward query-irrelevant visual cues, yielding unnecessary and even sometimes harmful reasoning traces (see~\cref{section:bottleneck}).
Furthermore, we observe a strong positive correlation between visual attention mass and LMMs’ effective visual activation, which motivates a simple yet effective visual mass guided budgeting strategy. 
Specifically, VisionPulse uses this lightweight mass signal to set a per-step retention budget and retains only the most critical visual tokens under this budget.
Extensive experiments on seven benchmarks demonstrate that VisionPulse achieves performance comparable to the full-token baseline under extreme visual token retention settings (10\% and 5\%), while reducing unnecessary generation length by 12.3\% and 11.2\%, substantially outperforming existing methods.

Our contributions are summarized as follows:
\textbf{1)} We empirically show that the need for visual evidence in multimodal reasoning is highly step-dependent, and the critical visual token set evolves over decoding steps.
\textbf{2)} We identify a coupled bottleneck in multimodal reasoning, where redundant visual context can steer LMMs to query-irrelevant regions and lengthen the reasoning trace.
\textbf{3)} We propose \textbf{VisionPulse}, a training-free dynamic visual token pruning framework that selects critical visual tokens at each decoding step via a lightweight visual mass signal.
\textbf{4)} Extensive experiments on seven benchmarks validate that VisionPulse mitigates redundant reasoning traces, improving efficiency while keeping accuracy almost unchanged.

\vspace{-0.02in}
\section{Related Work}

\textbf{Multimodal Reasoning}~~
Large Reasoning Models~\cite{deepseekr1,openaio1,team2025kimi} improve performance by explicitly generating chain-of-thought (CoT), but stronger reasoning typically requires longer generated sequences and thus higher inference cost, especially on difficult problems~\cite{muennighoff2025s1,snell2024scaling,yeo2025demystifying}. 
Efficient reasoning methods~\cite{thinkclearly,xia2025tokenskip,adaptthink,fastslow,yang2025lookback} therefore primary focus on more concise CoT generation while largely ignoring the contribution of visual tokens to inference cost.
In contrast, we first study multimodal reasoning efficiency from a unified perspective and show that redundant visual context is doubly costly: it increases per-step attention computation and can also lengthen the reasoning trace.

\textbf{Visual Compression}~~
Visual token redundancy is prevalent in LMMs: a single high-resolution image can yield thousands of tokens, and videos can scale to tens of thousands~\cite{hrbench,longvideobench}. 
Existing pruning methods broadly fall into two categories:
(1) compressing visual tokens using attention patterns from the vision encoder~\cite{visionzip,vflowopt,prumerge,arif2025hired,liu2025hiprune}, and
(2) selecting a query-relevant subset at prefill~\cite{fastv,sparsevlm,sparsevila,zhang2025beyond}.
Despite these differences, most methods compress only once before decoding, implicitly assuming the retained subset remains sufficient throughout reasoning.
In contrast, our analysis shows that visual reliance varies substantially across decoding steps, which directly motivates the design of VisionPulse.

\vspace{-0.02in}

\vspace{-0.02in}
\section{Motivation}

\subsection{Inference Cost Analysis of Multimodal Reasoning}

In multimodal reasoning, given a test query $q = \{p, v\}$, where $p$ is the text prompt and $v$ is the visual input (images or videos), LMMs integrate visual and textual information to generate an answer through multi-step reasoning.
Under the standard KV-cache implementation, the model first processes the full input once to initialize the KV-cache (prefill stage), and then generates $g$ tokens autoregressively while updating the KV-cache (decoding stage). 
We decompose the total floating-point operations (FLOPs) as follows:
\begin{equation}
\begin{aligned}
\mathcal{F}_{\text{total}} &= \mathcal{F}_{\text{pre}} + \sum_{t=1}^{g} \mathcal{F}_{\text{dec}} \\
&\approx L \cdot \Big[ \underbrace{(p+v)(8d^2 + 4md) + 4d(p+v)^2}_{\text{Prefill Stage}} \Big] \\
&\quad + L \cdot \underbrace{\sum_{t=1}^{g}\Big[ (8d^2 + 4md) + 4d(p+v+t) \Big]}_{\text{Decoding Stage}},
\end{aligned}
\label{eq:flops_analysis}
\end{equation}
where $L$ denotes the number of layers, $d$ is the hidden dimension, $m$ is the FFN intermediate size, and $g$ is the number of generated tokens. 
Eq.~\ref{eq:flops_analysis} highlights a coupling inference burden between the reasoning length $g$ and the initial multimodal context length $(p+v)$.
Unlike prefill which is performed once, decoding applies self-attention over a context growing from $(p+v)$ to $(p+v+g)$, the total cost scales as $O\!\left(g(p+v)+g^2\right)$.
In multimodal settings where the visual context often dominates the input length ($v \gg p$), visual tokens introduce a large baseline cost per decoding step.
As $g$ increase, this results in significantly higher marginal costs compared to text-only models, making the visual context becoming a dominant factor in total end-to-end latency.

\vspace{-0.04in}

\subsection{Dynamic Visual Attention in Multimodal Reasoning}

Given the redundancy in visual tokens, visual compression has emerged as a promising strategy for enhancing efficiency.
Most existing methods adopt a static selection paradigm: they estimate token importance once at prefill and keep this fixed subset throughout decoding.
This implicitly assumes that the visual evidence required for reasoning remains unchanged over time.
We argue that this assumption is mismatched to multimodal reasoning, where the model’s reliance on visual evidence changes as the reasoning state evolves.
To validate this, we visualize visual attention in Qwen3-VL-4B-Thinking during CoT reasoning as shown in~\cref{figure_observation}.
Specifically, we quantify step-wise visual attention mass and examine visual attention heatmaps at representative steps, leading to three key observations.

First, visual reliance is strongly step-dependent rather than constant.
As shown in \cref{figure_observation}a, visual attention activates when LMM confirms visual details (such as \emph{handle}), while diminishing significantly during steps dominated by textual reasoning (e.g., \emph{``Let's check the options"}).
It indicate that visual evidence is dynamic and activated on demand as the reasoning state evolves.
Second, the set of critical visual tokens varies during reasoning.
When visual attention mass increases, the activated regions also shift with the current reasoning content.
In \cref{figure_observation}b, when the model generates entity-specific tokens (e.g., \emph{handle}), attention explicitly concentrates on the corresponding region.
In contrast, during prefill, the model focuses on generating the initial token (e.g., \emph{so} in step 1), providing little semantic guidance for identifying the relevant regions.
Consequently, selecting a single fixed visual subset at prefill is suboptimal: it may discard tokens that become important later, while retaining irrelevant context during low-reliance steps.

Third and critically, we identify a coupled bottleneck in multimodal reasoning.
Maintaining the full visual context throughout decoding can bias the model toward generating unnecessary visual cues that contribute little to the reasoning process.
For instance, in~\cref{figure_observation}, describing \emph{handle} is irrelevant for judging the spatial relation between the door and the brown square.
Such redundancy not only incurs avoidable computational overhead but also introduces visual noise that undermines the reliability of the reasoning.
We provide a more comprehensive discussion of this phenomenon in \cref{section:bottleneck}.
Overall, these observations motivate a dynamic pruning mechanism that aligns visual budgets with evolving reasoning demands, maximizing compression efficiency without compromising LMMs' reasoning performance.

\section{VisionPulse}

\subsection{Overview Architecture}
\begin{figure}[t]
  % \vskip -0.1in
  \begin{center}
    \centerline{\includegraphics[width=\linewidth]{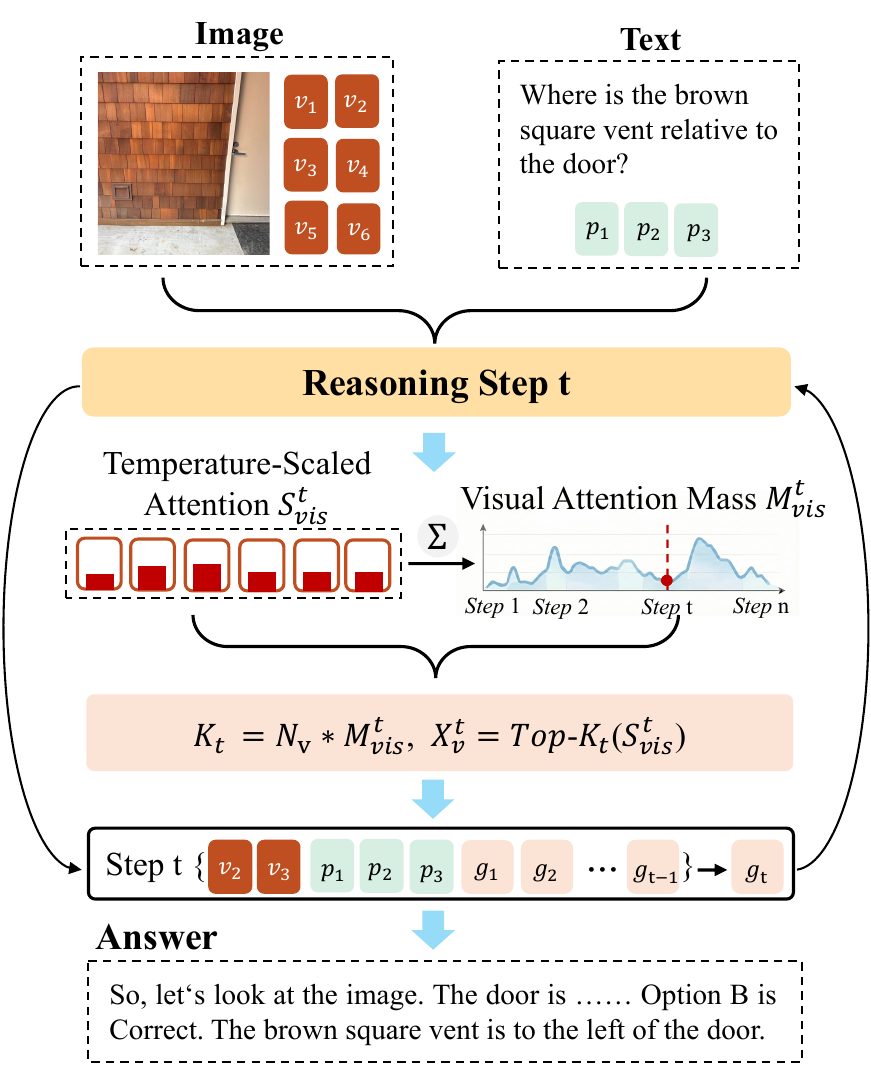}}
    \vskip-0.05in
    \caption{\textbf{Overview of VisionPulse.} At decoding step $t$, VisionPulse adaptively prunes visual tokens by computing a lightweight visual attention mass $M_{\mathrm{vis}}^{t}$ to determine the step-wise budget $K_t$, and retains the top-$K_t$ tokens for decoding. 
    Temperature scaling is used to enable adjustable compression ratios.
    }

    \label{fig:method}
  \end{center}
  \vskip-0.35in
\end{figure}

To mitigate the decoding-time bottleneck from redundant visual context, we propose \textbf{VisionPulse}, a \textbf{training-free} method for step-wise visual token pruning in multimodal reasoning.
As shown in \cref{fig:method}, VisionPulse keeps only the most critical visual tokens at each step, enabling step-dependent selection during decoding.
Furthermore, it uses a lightweight visual attention mass signal $M_{\mathrm{vis}}^{t}$ to determine the step-wise retention budget $K_t$, retaining the top-$K_t$ visual tokens to form the pruned decoding context.
We describe the details of each module below.

\vspace{-0.02in}
\subsection{Step-wise Visual Token Pruning}

In multimodal reasoning, LMMs exhibit highly step-dependent visual dependencies during decoding rather than maintaining a constant focus.
As a result, retaining the complete visual context at every decoding step imposes an unnecessary computation burden.
Motivated by this, we adopt a step-wise pruning paradigm that performs adaptive pruning at every decoding step.

Formally, let the visual token set be $X_v = \{v_1, v_2, ..., v_N\}$. 
At each decoding step $t$, given the current query token $q_t$, our objective is to identify a critical subset $X_v^t \subseteq X_v$ and drop the remaining tokens.
This fundamentally differs from static pruning methods, which estimate visual token importance once at prefill and keep a fixed subset throughout decoding.
In contrast, we re-estimate token importance at every step to track the evolving reasoning focus.
For simplicity, we follow FastV~\cite{fastv} for importance estimation but recompute it at every decoding step $t$.
Specifically, we compute the importance scores at layer $l_a$ and start pruning from $l_a$.
At the anchor layer $l_a$, we aggregate attention probabilities over all heads to define the significance score for each visual token:
\vspace{-2pt} % 减少上方间距
\begin{equation}
S_i^t \;=\; \frac{1}{H}\sum_{h=1}^{H} A^{(l_a)}_{t,h}\!\left(q_t, v_i\right),
\end{equation}
% \vspace{-2pt} % 减少下方间距
where $A^{(l_a)}_{t,h}(q_t, v_i)$ denotes the attention probability from the current query $q_t$ to the visual key $v_i$ at head $h$ of layer $l_a$.
This importance estimator is flexible and can be replaced by alternative scoring functions used in prior work~\cite{pyramiddrop,lifting}.
Notably, given $\{S_i^t\}_{i=1}^{N}$, instead of using a fixed retention ratio, we determine a step-specific retention budget $K_t$ and retain $\text{Top-}K_t$ visual tokens:
\begin{equation}
X_v^t \;=\; \{v_i \mid i \in \text{Top-}K_t(\{S_i^t\}_{i=1}^{N})\}.
\end{equation}

\subsection{Visual Mass Guided Dynamic Budget}

A key challenge in step-wise visual token pruning is that a fixed retention ratio cannot accommodate the heterogeneous visual needs across decoding steps.
Some steps require broad visual evidence, while others are largely language-driven and only need minimal visual support.
Furthermore, a larger visual attention does not necessarily imply a broader visual coverage: it could either concentrate on a few salient visual tokens or spread over many tokens.
To motivate a step-wise budget, we first conduct an empirical analysis across decoding steps and find that the visual attention mass $M_{\mathrm{vis}}^{t}$---a scalar measuring how much attention probability is allocated to visual tokens at step $t$---is strongly correlated with the number of effectively activated visual tokens.

\begin{figure}[t]
    \centering
    \centerline{\includegraphics[width=\linewidth]{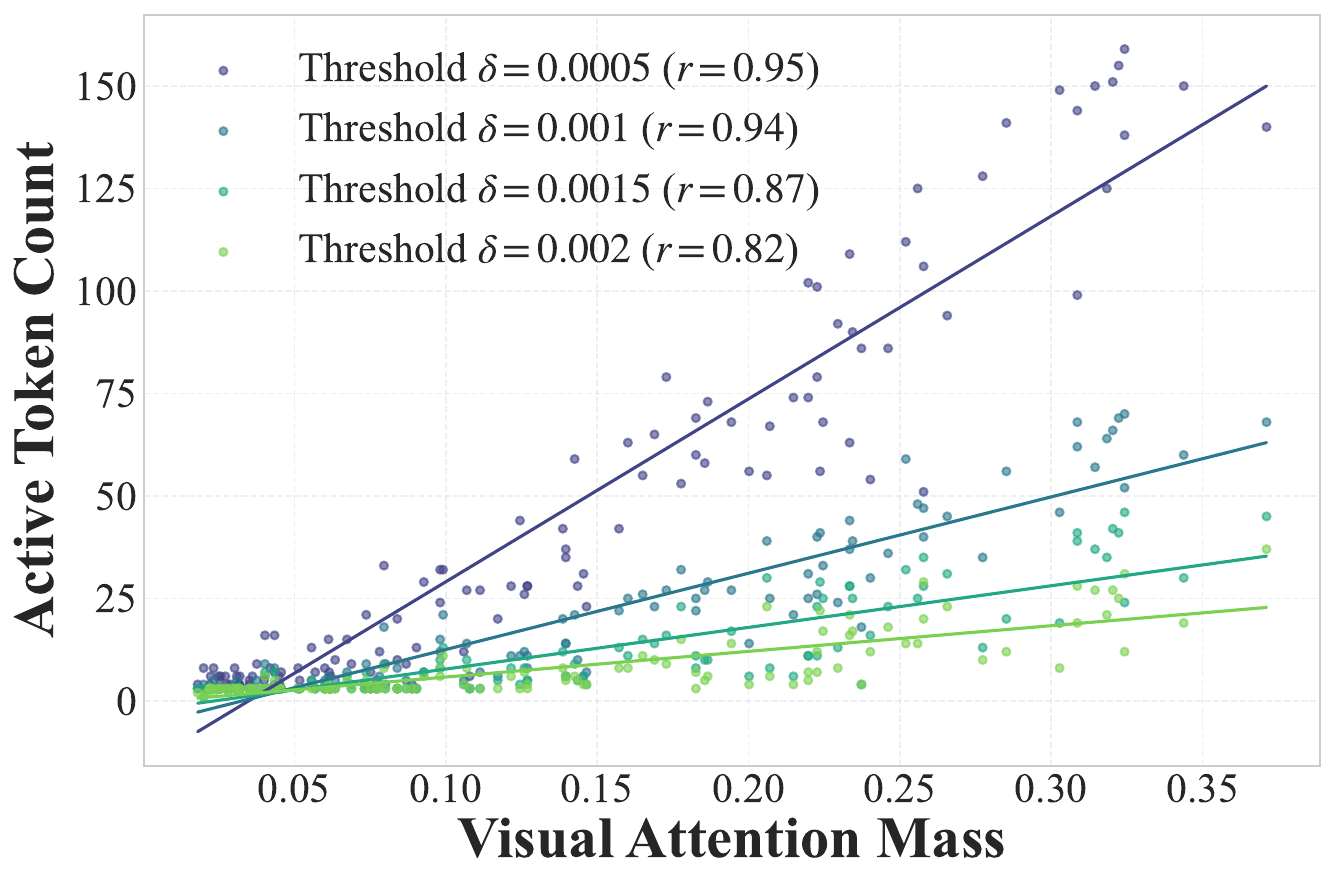}}
    \vspace{-0.08in}
    \caption{\textbf{Visual attention mass predicts the number of activated visual tokens.}
    Scatter plot of visual attention mass $M_{\mathrm{vis}}^{t}$ v.s. active visual token count $N_{\mathrm{act}}^{t}(\delta)$ under different activation thresholds $\delta$.
    Across thresholds, $M_{\mathrm{vis}}^{t}$ shows a strong positive linear correlation with $N_{\mathrm{act}}^{t}(\delta)$, supporting dynamic budget allocation.}
    \label{fig2:mass_vs_active}
    \vspace{-0.2in}
\end{figure}

Concretely, to characterize how many visual tokens receive non-negligible attention, we define the active visual token count under a threshold $\delta$ as:
\begin{equation}
\begin{aligned}
N_{\mathrm{act}}^{t}(\delta)
=\left|\left\{ i\in\{1,\ldots,N_v\} \;\middle|\; \bar{A}^{(l_a)}_{t}\!\left(q_t,v_i\right)>\delta \right\}\right|,
\end{aligned}
\end{equation}
where $N_v=|X_v|$ is the number of visual tokens.
Meanwhile, we quantify the overall visual dependency at decoding step $t$ by the visual attention mass:
\begin{equation}
%\small
m_{t,h}^{\mathrm{vis}} \;=\; \sum_{i=1}^{N_v} A^{(l_a)}_{t,h}\!\left(q_t, v_i\right),
\quad M_{\mathrm{vis}}^{t} \;=\; \frac{1}{H}\sum_{h=1}^{H} m_{t,h}^{\mathrm{vis}}. \label{eq:mass_visual} 
\end{equation}
As shown in~\cref{fig2:mass_vs_active}, $M_{\mathrm{vis}}^{t}$ exhibits a strong positive linear correlation with $N_{\mathrm{act}}^{t}(\delta)$ across decoding steps, and the trend remains consistent across a range of thresholds (Pearson $r$ ranges from $0.82$ to $0.95$).
This suggests that higher visual dependency tends to lift more tokens above non-negligible attention levels, instead of only amplifying attention on a few tokens.

Motivated by this observation, we use the visual attention mass as a lightweight predictor to allocate a step-wise retention budget.
At decoding step $t$, we estimate $M_{\mathrm{vis}}^{t}$ at the anchor layer $l_a$ using a temperature-scaled attention distribution with $\tau<1$, which enables controllable pruning ratios and suppresses the softmax long tail over $X_v$.
Specifically, we compute the temperature-scaled attention probability:
\begin{equation}
A^{(l_a,\tau)}_{t,h}(q_t, v_i)
=\frac{\exp\!\left(Z^{(l_a)}_{t,h}(q_t, v_i)/\tau\right)}
{\sum_{j}\exp\!\left(Z^{(l_a)}_{t,h}(q_t, j)/\tau\right)},
\label{eq6}
\end{equation}
where $Z^{(l_a)}_{t,h}(q_t, v_i)$ denotes the pre-softmax attention logit and $\tau$ is the temperature.
Using $A^{(l_a,\tau)}_{t,h}$, we compute the visual attention mass $M_{\mathrm{vis}}^{t}$ following \cref{eq:mass_visual}.
For budgeting, we adopt the head-max variant $M_{vis,max}^t$ when computing $K_t$, yielding a conservative estimate that avoids under-budgeting visual-dominant heads~\cite{kang2025see,wan2025only}.
Finally, we predict the pruning budget directly from $M_{\mathrm{vis,max}}^{t}$:
\begin{equation}
K_t \;=\; M_{\mathrm{vis,max}}^{t}\, N_v.
\end{equation}

% , and $\mathrm{clip}(\cdot)$ enforces minimum and maximum budgets

\newcommand{\rescell}[3]{%
  \makecell[c]{%
    \rule{0pt}{2.8ex}#1\\[-0.2ex]
    \textcolor{#3}{#2}%
  }%
}
\newcommand{\basecell}[1]{\makecell[c]{#1 \\ -}}
\newcommand{\baseone}[1]{\makecell[c]{#1}}
% =========================================================================
\vspace{-0.02in}
\begin{table*}[t]
\centering
\caption{\textbf{Main results under aggressive visual-token retention (Qwen3-VL Thinking 4B).} We compare our FastV-VP with prior pruning methods across seven benchmarks. Each cell reports the absolute value (top; $L$ = generated tokens, $A$ = accuracy) and the relative change from the full-token baseline (bottom). Results are grouped by retention levels (retain $\le$10\% and $\le$5\% tokens).}
\label{tab:main_results_qwen3vl_4b}
\vspace{-0.08in}

% Colors
\definecolor{badred}{RGB}{200, 0, 0}
\definecolor{goodgreen}{RGB}{0, 150, 0}
\definecolor{sectiongray}{RGB}{240, 240, 240}

% Table settings
\renewcommand{\arraystretch}{1.4}
\setlength{\tabcolsep}{1.3pt}

\scalebox{0.82}{
\begin{tabular}{l | c c | c c | c c | c c | c c | c c | c c | c c }
\toprule[1.5pt]

\multirow{2}{*}{\textbf{Method}} &
\multicolumn{2}{c|}{\textbf{CharXiv RQ}} &
\multicolumn{2}{c|}{\textbf{InfoVQA}} &
\multicolumn{2}{c|}{\textbf{ChartQA}} &
\multicolumn{2}{c|}{\textbf{MMStar}} &
\multicolumn{2}{c|}{\textbf{RealWorld QA}} &
\multicolumn{2}{c|}{\textbf{MMVet}} &
\multicolumn{2}{c|}{\textbf{MIA-Bench}} &
\multicolumn{2}{c}{\textbf{Avg}} \\
\cmidrule(lr){2-3}\cmidrule(lr){4-5}\cmidrule(lr){6-7}\cmidrule(lr){8-9}
\cmidrule(lr){10-11}\cmidrule(lr){12-13}\cmidrule(lr){14-15}\cmidrule(lr){16-17}
& $L$ $\downarrow$ & $A$ $\uparrow$ &
  $L$ $\downarrow$ & $A$ $\uparrow$ &
  $L$ $\downarrow$ & $A$ $\uparrow$ &
  $L$ $\downarrow$ & $A$ $\uparrow$ &
  $L$ $\downarrow$ & $A$ $\uparrow$ &
  $L$ $\downarrow$ & $A$ $\uparrow$ &
  $L$ $\downarrow$ & $A$ $\uparrow$ &
  $L$ $\downarrow$ & $A$ $\uparrow$ \\

% ===================================================
% Section 1: Baseline (100%)
% ===================================================
\specialrule{0.6pt}{0pt}{0pt}
\rowcolor{sectiongray}
\multicolumn{17}{c}{\textbf{Upper Bound (Retain 100\% Tokens)}} \\
\specialrule{0.6pt}{0pt}{0pt}

\textbf{Baseline} &
\baseone{4068.0} & \baseone{47.60} &
\baseone{623.1}  & \baseone{84.37} &
\baseone{510.0}  & \baseone{77.12} &
\baseone{2058.5} & \baseone{72.47} &
\baseone{678.0}  & \baseone{72.81} &
\baseone{1642.0} & \baseone{60.96} &
\baseone{2231.3} & \baseone{93.44} &
\baseone{1687.3} &
\baseone{72.68} \\

% ===================================================
% Section 2: 10% Ratio
% ===================================================
\specialrule{0.6pt}{0pt}{0pt}
\rowcolor{sectiongray}
\multicolumn{17}{c}{\textbf{Retain $\le$ 10\% Tokens}} \\
\specialrule{0.6pt}{0pt}{0pt}

Visionzip &
\rescell{4986.2}{+22.6\%}{badred} & \rescell{13.90}{-70.8\%}{badred} &
\rescell{2533.3}{+306.6\%}{badred} & \rescell{22.66}{-73.1\%}{badred} &
\rescell{2039.7}{+299.9\%}{badred} & \rescell{30.24}{-60.8\%}{badred} &
\rescell{2496.8}{+21.3\%}{badred} & \rescell{51.53}{-28.9\%}{badred} &
\rescell{726.0}{+7.1\%}{badred}  & \rescell{59.08}{-18.9\%}{badred} &
\rescell{2346.7}{+42.9\%}{badred} & \rescell{41.19}{-32.4\%}{badred} &
\rescell{3089.6}{+38.5\%}{badred} & \rescell{88.14}{-5.7\%}{badred} &
\rescell{2602.6}{+54.2\%}{badred} &
\rescell{43.82}{-39.7\%}{badred} \\

FastV &
\rescell{5960.1}{+46.5\%}{badred} & \rescell{12.70}{-73.3\%}{badred} &
\rescell{2963.6}{+375.6\%}{badred} & \rescell{20.63}{-75.5\%}{badred} &
\rescell{1485.5}{+191.3\%}{badred} & \rescell{16.28}{-78.9\%}{badred} &
\rescell{2275.3}{+10.5\%}{badred} & \rescell{50.40}{-30.5\%}{badred} &
\rescell{575.8}{-15.1\%}{goodgreen} & \rescell{55.95}{-23.2\%}{badred} &
\rescell{3296.8}{+100.8\%}{badred} & \rescell{28.12}{-53.9\%}{badred} &
\rescell{2714.9}{+21.7\%}{badred} & \rescell{82.53}{-11.7\%}{badred} &
\rescell{2753.1}{+63.2\%}{badred} &
\rescell{38.09}{-47.6\%}{badred} \\

LOOK-M &
\rescell{5555.2}{+36.6\%}{badred} & \rescell{19.80}{-58.4\%}{badred} &
\rescell{2694.1}{+332.4\%}{badred} & \rescell{40.94}{-51.5\%}{badred} &
\rescell{2007.1}{+293.5\%}{badred} & \rescell{57.68}{-25.2\%}{badred} &
\rescell{2090.6}{+1.6\%}{badred}  & \rescell{58.87}{-18.8\%}{badred} &
\rescell{481.8}{-28.9\%}{goodgreen} & \rescell{67.84}{-6.8\%}{badred} &
\rescell{2518.9}{+53.4\%}{badred} & \rescell{50.05}{-17.9\%}{badred} &
\rescell{3376.1}{+51.3\%}{badred} & \rescell{88.71}{-5.1\%}{badred} &
\rescell{2674.8}{+58.5\%}{badred} &
\rescell{54.84}{-24.5\%}{badred} \\

\makecell[l]{\textbf{FastV-VP}\\\textbf{(Ours)}} &
\rescell{3770.7}{-7.3\%}{goodgreen} & \rescell{47.30}{-0.6\%}{badred} &
\rescell{530.7}{-14.8\%}{goodgreen} & \rescell{83.62}{-0.9\%}{badred} &
\rescell{422.9}{-17.1\%}{goodgreen} & \rescell{76.72}{-0.5\%}{badred} &
\rescell{1873.7}{-9.0\%}{goodgreen} & \rescell{72.20}{-0.4\%}{badred} &
\rescell{564.0}{-16.8\%}{goodgreen} & \rescell{73.20}{+0.5\%}{goodgreen} &
\rescell{1329.8}{-19.0\%}{goodgreen} & \rescell{61.79}{+1.4\%}{goodgreen} &
\rescell{1870.8}{-16.2\%}{goodgreen} & \rescell{93.99}{+0.6\%}{goodgreen} &
\rescell{1480.4}{-12.3\%}{goodgreen} &
\rescell{72.69}{+0.0\%}{goodgreen} \\

% ===================================================
% Section 3: 5% Ratio
% ===================================================
\specialrule{0.6pt}{0pt}{0pt}
\rowcolor{sectiongray}
\multicolumn{17}{c}{\textbf{Retain $\le$ 5\% Tokens}} \\
\specialrule{0.6pt}{0pt}{0pt}

Visionzip &
\rescell{3751.7}{-7.8\%}{goodgreen} & \rescell{12.30}{-74.2\%}{badred} &
\rescell{1468.5}{+135.7\%}{badred} & \rescell{22.04}{-73.9\%}{badred} &
\rescell{957.3}{+87.7\%}{badred} & \rescell{19.43}{-74.8\%}{badred} &
\rescell{2289.6}{+11.2\%}{badred} & \rescell{44.93}{-38.0\%}{badred} &
\rescell{671.2}{-1.0\%}{goodgreen} & \rescell{52.02}{-28.6\%}{badred} &
\rescell{1918.1}{+16.8\%}{badred} & \rescell{36.56}{-40.0\%}{badred} &
\rescell{2914.1}{+30.6\%}{badred} & \rescell{87.88}{-6.0\%}{badred} &
\rescell{1995.8}{+18.3\%}{badred} &
\rescell{39.31}{-45.9\%}{badred} \\

FastV &
\rescell{4259.7}{+4.7\%}{badred} & \rescell{11.60}{-75.6\%}{badred} &
\rescell{2239.7}{+259.4\%}{badred} & \rescell{21.78}{-74.2\%}{badred} &
\rescell{1207.3}{+136.7\%}{badred} & \rescell{12.88}{-83.3\%}{badred} &
\rescell{2040.4}{-0.9\%}{goodgreen} & \rescell{50.13}{-30.8\%}{badred} &
\rescell{613.3}{-9.5\%}{goodgreen} & \rescell{54.12}{-25.7\%}{badred} &
\rescell{2606.5}{+58.7\%}{badred} & \rescell{24.27}{-60.2\%}{badred} &
\rescell{2342.2}{+5.0\%}{badred} & \rescell{75.03}{-19.7\%}{badred} &
\rescell{2187.0}{+29.6\%}{badred} &
\rescell{35.69}{-50.9\%}{badred} \\

LOOK-M &
\rescell{6448.0}{+58.5\%}{badred} & \rescell{17.20}{-63.9\%}{badred} &
\rescell{5511.1}{+784.5\%}{badred} & \rescell{27.39}{-67.5\%}{badred} &
\rescell{2577.2}{+405.3\%}{badred} & \rescell{28.80}{-62.7\%}{badred} &
\rescell{2265.4}{+10.1\%}{badred} & \rescell{54.87}{-24.3\%}{badred} &
\rescell{573.0}{-15.5\%}{goodgreen} & \rescell{68.37}{-6.1\%}{badred} &
\rescell{2859.2}{+74.1\%}{badred} & \rescell{35.05}{-42.5\%}{badred} &
\rescell{4329.3}{+94.0\%}{badred} & \rescell{80.66}{-13.7\%}{badred} &
\rescell{3509.0}{+108.0\%}{badred} &
\rescell{44.62}{-38.6\%}{badred} \\

\makecell[l]{\textbf{FastV-VP}\\\textbf{(Ours)}} &
\rescell{3645.1}{-10.4\%}{goodgreen} & \rescell{45.20}{-5.0\%}{badred} &
\rescell{665.0}{+6.7\%}{badred} & \rescell{81.90}{-2.9\%}{badred} &
\rescell{510.0}{+0.0\%}{goodgreen} & \rescell{75.16}{-2.5\%}{badred} &
\rescell{1875.4}{-8.9\%}{goodgreen} & \rescell{70.53}{-2.7\%}{badred} &
\rescell{538.0}{-20.6\%}{goodgreen} & \rescell{72.54}{-0.4\%}{badred} &
\rescell{1456.2}{-11.3\%}{goodgreen} & \rescell{59.00}{-3.2\%}{badred} &
\rescell{1800.1}{-19.3\%}{goodgreen} & \rescell{95.09}{+1.8\%}{goodgreen} &
\rescell{1498.5}{-11.2\%}{goodgreen} &
\rescell{71.35}{-1.8\%}{badred} \\

\bottomrule[1.5pt]
\end{tabular}
}
\label{tab:main_results}
\vspace{-0.1in}
\end{table*}

\vspace{-0.08in}
\section{Experiments}

\subsection{Experimental Setup}

To evaluate our method on general multimodal reasoning, we conduct experiments on seven widely adopted benchmarks~\cite{chartqa,mmvet,realworldqa,miabench,charxiv,infovqa,mmstar}.
We compare our approach with three representative state-of-the-art visual token pruning methods: VisionZip~\cite{visionzip}, FastV~\cite{fastv}, and LOOK-M~\cite{lookm}.
VisionZip selects visual tokens according to the visual attention extracted from the last layer of the visual encoder, and performs pruning before the prefill stage.
FastV prunes visual tokens at an intermediate layer of the LMM during the prefill stage, where $l_a =17$.
LOOK-M performs KV-cache compression across all layers during prefill, preserving important tokens in a layer-wise manner.
Despite differing in pruning granularity and application stage, these methods make a single static pruning decision and therefore overlook the step-dependent need for visual evidence during reasoning.
For a controlled comparison, we build VisionPulse on top of FastV’s importance estimator (FastV-VP) and prune at the same layer.

\subsection{Main Results}

We evaluate all methods under two aggressive settings with $\sim$10\% and $\sim$5\% visual token retention.
For fair comparison, we clamp $K_t$ to match the target average retention: $(0.05N, 0.10N)$ for 10\% and $(0.01N, 0.05N)$ for 5\%.

The results are shown in \cref{tab:main_results}, where our method consistently achieves markedly better performance than existing approaches.
Specifically, when using at most 5\% visual tokens at each decoding step, our method preserves nearly 100\% of the original performance. 
Even under the more extreme setting where only 1\% of visual tokens are retained per step, the model still maintains 98.2\% of the full-token performance.
Moreover, unlike prior methods, we significantly shorten the reasoning trajectory while keeping the performance nearly unchanged. 
Compared to the full-token baseline, the average generation length is reduced by 12.3\% and 11.2\% under the two settings, respectively. 
In contrast, static pruning baselines degrade sharply, with average accuracy dropping by 24.5\%--50.9\% and generation length increasing by 18.3\%--108.0\% under aggressive retention.
These results indicate that VisionPulse does not only reduce the number of visual tokens, but instead removes truly irrelevant visual information at each decoding step, consistently retaining only the tokens critical to sustain effective visual activation for the current reasoning stage.

Furthermore, we observe that incorrect pruning strategies not only degrade performance but also unexpectedly increase reasoning cost substantially.
Static methods (e.g., VisionZip, FastV, and LOOK-M) often remove critical tokens prematurely, forcing the model to compensate via longer reasoning generation, which both reduces accuracy and increases computational overhead.
For example, LOOK-M incurs a 108\% increase in average generation length, yet its accuracy still drops by 38.6\%.
In contrast, VisionPulse maintains shorter generations and stable performance even under extremely low visual retention. 
Even on vision-intensive tasks such as ChartQA, our method remains stable, with only a 2.5\% performance drop under 5\% visual-token retention.
Overall, these results support our hypothesis that visual dependency varies throughout the reasoning process, and fixed-budget pruning strategies are inherently mismatched to this property.
By using visual attention mass as a lightweight predictor, VisionPulse better tracks step-wise visual dependency and enables more efficient pruning.

\begin{figure*}[t]   
    \centering
    \centerline{\includegraphics[width=\linewidth]{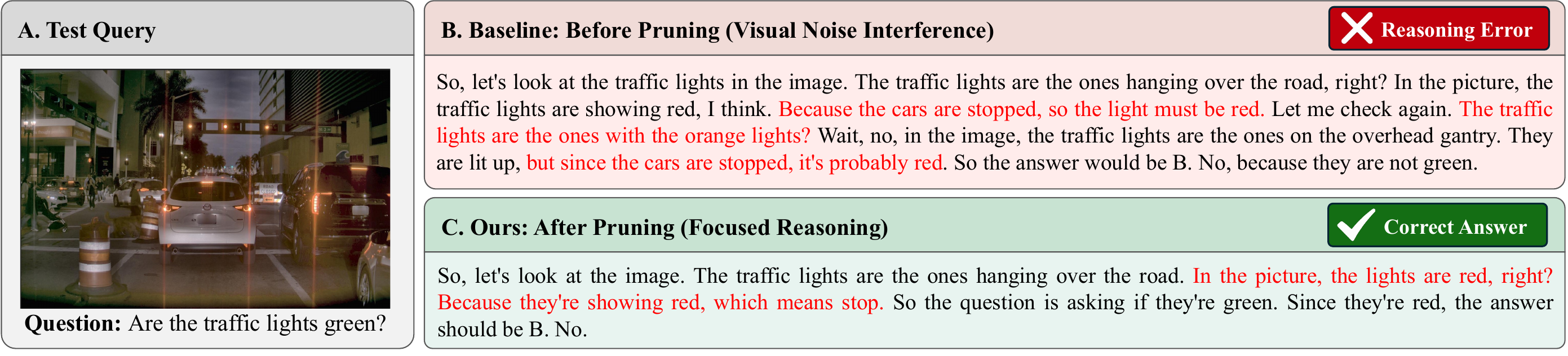}}
    \vspace{-0.08in}
    \caption{\textbf{Coupled bottleneck in multimodal reasoning.}
    With full visual context (top), redundant visual tokens remain available throughout decoding and can draw attention to query-irrelevant cues, leading to unnecessary descriptions and even erroneous reasoning (e.g., inferring the traffic-light state from vehicles rather than the signal). 
    With step-wise visual pruning (bottom), the model retains only query-relevant visual evidence for the current step, yielding more focused reasoning and correct prediction.}
  \label{fig3:mass_vs_active}  
  \vspace{-0.0in}
\end{figure*}

\begin{table*}[t]
\centering
\caption{\textbf{Ablation study on Qwen3-VL-Thinking-4B under aggressive retention ($\le$5\%).} We analyze the effect of different budget allocation strategies used for FastV-VP on three benchmarks. }
\label{tab:ablation}
\vspace{-0.08in}

\definecolor{badred}{RGB}{200, 0, 0}
\definecolor{goodgreen}{RGB}{0, 150, 0}
\definecolor{sectiongray}{RGB}{240, 240, 240}

\renewcommand{\arraystretch}{1.2}
\setlength{\tabcolsep}{10.5pt}
\scalebox{0.85}{
\begin{tabular}{l | c | c c | c c | c c | c c}
\toprule[1.5pt]
\multirow{2}{*}{\textbf{Method}} &
\multirow{2}{*}{\textbf{Avg Ret.}} &
\multicolumn{2}{c|}{\textbf{RealWorld QA}} &
\multicolumn{2}{c|}{\textbf{MMVet}} &
\multicolumn{2}{c|}{\textbf{MIA-Bench}} &
\multirow{2}{*}{\textbf{Avg $L$} $\downarrow$} &
\multirow{2}{*}{\textbf{Avg $A$} $\uparrow$} \\
\cmidrule(lr){3-4}\cmidrule(lr){5-6}\cmidrule(lr){7-8}
& &
$L$ $\downarrow$ & $A$ $\uparrow$ &
$L$ $\downarrow$ & $A$ $\uparrow$ &
$L$ $\downarrow$ & $A$ $\uparrow$ &
& \\

% ===================================================
% Baseline
% ===================================================
% \specialrule{0.6pt}{0pt}{0pt}
% \rowcolor{sectiongray}
% \multicolumn{10}{c}{\textbf{Upper Bound (Retain 100\% Tokens)}} \\
% \specialrule{0.6pt}{0pt}{0pt}
\midrule
\textbf{Baseline} &
100\% &
\baseone{678.0} & \baseone{72.81} &
\baseone{1642.0} & \baseone{60.96} &
\baseone{2231.3} & \baseone{93.44} &
% \rescell{1517.1}{0.0\%}{goodgreen} &
% \rescell{75.74}{0.0\%}{goodgreen} \\
\baseone{1517.1}    &
\baseone{75.74}     \\

% ===================================================
% <=5% regime
% ===================================================
\specialrule{0.6pt}{0pt}{0pt}
\rowcolor{sectiongray}
\multicolumn{10}{c}{\textbf{Retain $\le$ 5\% Tokens}} \\
\specialrule{0.6pt}{0pt}{0pt}

FastV &
5.0\% &
\rescell{613.3}{-9.5\%}{badred} & \rescell{54.12}{-25.7\%}{badred} &
\rescell{2606.5}{+58.7\%}{badred} & \rescell{24.27}{-60.2\%}{badred} &
\rescell{2342.2}{+5.0\%}{badred} & \rescell{75.03}{-19.7\%}{badred} &
\rescell{1853.9}{+22.2\%}{badred} &
\rescell{51.14}{-32.5\%}{badred} \\
\midrule
\makecell[l]{FastV-VP\\(Fix Ratio)} &
1.0\% &
\rescell{509.7}{-24.8\%}{goodgreen} & \rescell{71.90}{-1.3\%}{badred} &
\rescell{2913.4}{+77.4\%}{badred} & \rescell{49.17}{-19.3\%}{badred} &
\rescell{2398.0}{+7.5\%}{badred} & \rescell{92.03}{-1.5\%}{badred} &
\rescell{1940.4}{+27.9\%}{badred} &
\rescell{71.03}{-6.2\%}{badred} \\

\makecell[l]{FastV-VP\\(Fix Ratio)} &
5.0\% &
\rescell{678.6}{+0.1\%}{badred} & \rescell{72.81}{+0.0\%}{goodgreen} &
\rescell{1557.7}{-5.1\%}{goodgreen} & \rescell{59.45}{-2.5\%}{badred} &
\rescell{1971.0}{-11.7\%}{goodgreen} & \rescell{93.22}{-0.2\%}{badred} &
\rescell{1402.4}{-7.6\%}{goodgreen} &
\rescell{75.16}{-0.8\%}{badred} \\

\makecell[l]{FastV-VP\\(Random Budget)} &
3.0\% &
\rescell{519.0}{-23.5\%}{goodgreen} & \rescell{69.28}{-4.9\%}{badred} &
\rescell{1820.9}{+10.9\%}{badred} & \rescell{58.02}{-4.8\%}{badred} &
\rescell{2219.5}{-0.5\%}{goodgreen} & \rescell{91.49}{-2.1\%}{badred} &
\rescell{1519.8}{+0.2\%}{badred} &
\rescell{72.93}{-3.7\%}{badred} \\
\midrule
\makecell[l]{\textbf{FastV-VP}\\\textbf{(Dynamic Budget)}} &
1.9\% &
\rescell{538.0}{-20.6\%}{goodgreen} & \rescell{72.54}{-0.4\%}{badred} &
\rescell{1456.2}{-11.3\%}{goodgreen} & \rescell{59.00}{-3.2\%}{badred} &
\rescell{1800.1}{-19.3\%}{goodgreen} & \rescell{95.09}{+1.8\%}{goodgreen} &
\rescell{\textbf{1264.8}}{\textbf{-16.6\%}}{goodgreen} &
\rescell{\textbf{75.54}}{\textbf{-0.3\%}}{badred} \\

\bottomrule[1.5pt]
\end{tabular}
}
\vspace{-0.05in}
\end{table*}

\subsection{The Coupled Bottleneck in Multimodal Reasoning}
\label{section:bottleneck}
Beyond inference-time overhead, we observe a coupled bottleneck that affects both the length and reliability of multimodal reasoning.
During decoding, LMMs remain exposed to abundant visual context even when the current step requires little visual evidence, which can steer the model toward query-irrelevant regions and induce redundant descriptions, thereby elongating the reasoning trace.

As illustrated in \cref{figure_observation,fig3:mass_vs_active}, this phenomenon not only lengthens reasoning traces but can also lead the model toward harmful reasoning paths.
Specifically, in \cref{figure_observation}a, the model elaborates on descriptive details (e.g., the handle and color of the door) that are irrelevant to the target reasoning.
Furthermore, in \cref{fig3:mass_vs_active}, an early description of the car biases subsequent inference, prompting the model to infer the traffic light state from the car’s motion instead of the signal itself. 
Such failure modes are unacceptable in safety-critical scenarios such as autonomous driving. 
Additional examples are provided in the Appendix.
In contrast, step-wise visual sparsification retains only the evidence required at each step, reducing redundant generations while preserving correctness.
These observations support our claim that step-wise pruning paradigm better aligns the visual budget with evolving visual reliance during reasoning.

\subsection{Ablation Study}

We conduct ablation studies to investigate the contribution of each component in VisionPulse. 
As shown in \cref{tab:ablation}, both components of our method are effective and contribute to the overall gains.

(1) \textbf{Step-wise pruning is necessary:}
% extremely 
Under low retention, aligning pruning with the step-wise need for visual evidence is critical for both efficiency and correctness.
Using FastV as a static baseline, fixed 5\% retention leads to longer generations (+22.2\%) and a large accuracy drop (-32.5\%) in \cref{tab:ablation}.
This suggests that prefill-only pruning removes evidence required later, forcing the LMM to compensate with longer reasoning yet still failing to answer correctly.
(2) \textbf{Dynamic budgeting improves the trade-off:} within FastV-VP, a fixed 1\% budget causes a 6.2\% accuracy drop with 27.9\% longer generations, while a fixed 5\% budget cannot adapt to step-wise variation and leaves redundant context in low-demand steps.
In contrast, dynamic budgeting achieves a better balance, reducing length by 16.6\% with negligible accuracy change (0.3\%), and outperforms random budgeting in both length and accuracy.

\begin{table}[t]
\centering
\caption{\textbf{Generalization across LMM backbones.}
The results of VisionPulse on RealWorldQA and MIA-Bench for different LMMs.
FastV~\cite{fastv} is included for comparision.
}
\label{tab:generalization_5pct}
\vspace{-0.08in}

\definecolor{badred}{RGB}{200, 0, 0}
\definecolor{goodgreen}{RGB}{0, 150, 0}
\definecolor{sectiongray}{RGB}{240, 240, 240}

\renewcommand{\arraystretch}{1.35}
\setlength{\tabcolsep}{4.5pt}
\scalebox{0.80}{
\begin{tabular}{l | c c | c c | c | c}
\toprule[1.5pt]
\multirow{2}{*}{\textbf{Method}} &
\multicolumn{2}{c|}{\textbf{RealWorldQA}} &
\multicolumn{2}{c|}{\textbf{MIA-Bench}} &
\multirow{2}{*}{\textbf{Avg $L$} $\downarrow$} &
\multirow{2}{*}{\textbf{Avg $A$} $\uparrow$} \\
\cmidrule(lr){2-3}\cmidrule(lr){4-5}
& $L$ $\downarrow$ & $A$ $\uparrow$ &
  $L$ $\downarrow$ & $A$ $\uparrow$ & & \\
\specialrule{0.6pt}{0pt}{0pt}

% ===================== Qwen3-VL-Thinking 8B =====================
\rowcolor{sectiongray}
\multicolumn{7}{c}{\textbf{Qwen3-VL-Thinking 8B}} \\
\specialrule{0.6pt}{0pt}{0pt}

\textbf{Baseline} &
\baseone{408.4} & \baseone{77.24} &
\baseone{1325.2} & \baseone{94.19} &
\baseone{866.8} &
\baseone{85.72} \\
\specialrule{0.6pt}{0pt}{0pt}

\rowcolor{sectiongray}
\multicolumn{7}{c}{\textbf{Retain $\le$ 5\% Tokens}} \\
\specialrule{0.6pt}{0pt}{0pt}

\textbf{FastV} &
\rescell{348.1}{-14.8\%}{goodgreen} & \rescell{57.52}{-25.5\%}{badred} &
\rescell{1719.8}{+29.8\%}{badred} & \rescell{89.29}{-5.20\%}{badred} &
\rescell{1033.95}{+19.3\%}{badred} &
\rescell{73.41}{-14.4\%}{badred} \\

\textbf{FastV-VP} &
\rescell{400.0}{-2.1\%}{goodgreen} & \rescell{76.60}{-0.8\%}{badred} &
\rescell{1196.5}{-9.7\%}{goodgreen} & \rescell{95.13}{+1.0\%}{goodgreen} &
\rescell{798.2}{-7.9\%}{goodgreen} &
\rescell{85.87}{+0.2\%}{goodgreen} \\
\specialrule{0.6pt}{0pt}{0pt}

% ===================== InternVL 3.5-Thinking 4B =====================
\rowcolor{sectiongray}
\multicolumn{7}{c}{\textbf{InternVL-3.5-Thinking 4B}} \\
\specialrule{0.6pt}{0pt}{0pt}

\textbf{Baseline} &
\baseone{3707.5} & \baseone{54.38} &
\baseone{1756.9} & \baseone{89.70} &
\baseone{2732.2} &
\baseone{72.04} \\
\specialrule{0.6pt}{0pt}{0pt}

\rowcolor{sectiongray}
\multicolumn{7}{c}{\textbf{Retain $\le$ 5\% Tokens}} \\
\specialrule{0.6pt}{0pt}{0pt}

\textbf{FastV} &
\rescell{3787.9}{+2.2\%}{badred} & \rescell{44.18}{-18.8\%}{badred} &
\rescell{2274.5}{+29.5\%}{badred} & \rescell{82.63}{-7.9\%}{badred} &
\rescell{3031.2}{+11.0\%}{badred} &
\rescell{63.41}{-12.0\%}{badred} \\

\textbf{FastV-VP} &
\rescell{3246.5}{-12.4\%}{goodgreen} & \rescell{55.16}{+1.4\%}{goodgreen} &
\rescell{1537.2}{-12.5\%}{goodgreen} & \rescell{90.10}{+0.4\%}{goodgreen} &
\rescell{2391.9}{-12.4\%}{goodgreen} &
\rescell{72.63}{+0.8\%}{goodgreen} \\

\bottomrule[1.5pt]
\end{tabular}
}
\vspace{-0.3in}
\end{table}

\subsection{Generalization across different LMMs}

We evaluate LMMs from different families and at different scales to verify VisionPulse's generalization.
\cref{tab:generalization_5pct} demonstrates that VisionPulse generalizes across different LMM backbones under 5\% retention. 
On both Qwen3-VL-Thinking 8B and InternVL-3.5-Thinking 4B, FastV consistently degrades accuracy with increasing generation length (e.g., Avg $A$ drops by 14.4\% and 12.0\%, while Avg $L$ increases by 19.3\% and 11.0\%). 
In contrast, FastV-VP preserves accuracy while reducing generation length: it slightly improves Avg $A$ (+0.2/+0.8\%) with shorter Avg $L$ (-7.9/-12.4\%), indicating stable gains across different LMMs.

\subsection{Effect of Different Importance Estimators}

VisionPulse is a general framework for pruning during multimodal reasoning, emphasizing step-wise visual budget allocation rather than a specific token-importance estimator.
To validate this, we integrate VisionPulse with FastV~\cite{fastv} and Pdrop~\cite{pyramiddrop}, and consider an ablation variant (\textit{Ll}) that performs layer-wise dynamic pruning (instead of token-level selection) for a direct comparison.
We evaluate all three variants on RealWorldQA and MIA-Bench under two retention settings.

As shown in \cref{tab:importance_realworld_mia}, VisionPulse consistently improves the trade-off between efficiency and accuracy across different importance estimators.
Under the 5\% retention setting, FastV-VP achieves the best overall balance, reducing the average generation length by 19.6\% while slightly improving accuracy, whereas Pdrop-VP achieves larger length reductions but incurs a modest accuracy drop.
Notably, the layer-wise ablation Ll-VP achieves a comparable efficiency and accuracy trade-off as the token-level variant, suggesting that pushing the strategy to the layer level offers no clear advantage over token-level dynamic pruning.
Overall, these results demonstrate that VisionPulse is compatible with different importance estimators, and that the primary gains stem from token-level dynamic budgeting.

\begin{table}[t]
\centering
\caption{\textbf{Effect of different importance estimators.}
Results on RealWorldQA and MIA-Bench using different importance estimators within VisionPulse (VP).}
\label{tab:importance_realworld_mia}
\vspace{-0.08in}

\definecolor{badred}{RGB}{200, 0, 0}
\definecolor{goodgreen}{RGB}{0, 150, 0}
\definecolor{sectiongray}{RGB}{240, 240, 240}

\renewcommand{\arraystretch}{1.3}
\setlength{\tabcolsep}{5pt}
\scalebox{0.80}{
\begin{tabular}{l | c c | c c | c | c}
\toprule[1.5pt]
\multirow{2}{*}{\textbf{Method}} &
\multicolumn{2}{c|}{\textbf{RealWorldQA}} &
\multicolumn{2}{c|}{\textbf{MIA-Bench}} &
\multirow{2}{*}{\textbf{Avg $L$} $\downarrow$} &
\multirow{2}{*}{\textbf{Avg $A$} $\uparrow$} \\
\cmidrule(lr){2-3}\cmidrule(lr){4-5}
& $L$ $\downarrow$ & $A$ $\uparrow$ &
  $L$ $\downarrow$ & $A$ $\uparrow$ &
  & \\
\specialrule{0.6pt}{0pt}{0pt}

\rowcolor{sectiongray}
\multicolumn{7}{c}{\textbf{Upper Bound (Retain 100\% Tokens))}} \\
\specialrule{0.6pt}{0pt}{0pt}

\textbf{Baseline} &
\baseone{678.0} & \baseone{72.81} &
\baseone{2231.3} & \baseone{93.44} &
\baseone{1454.7} &
\baseone{83.12} \\
\specialrule{0.6pt}{0pt}{0pt}

\rowcolor{sectiongray}
\multicolumn{7}{c}{\textbf{Retain $\le$ 10\% Tokens}} \\
\specialrule{0.6pt}{0pt}{0pt}

FastV-VP &
\rescell{578.0}{-14.7\%}{goodgreen} & \rescell{73.20}{+0.5\%}{goodgreen} &
\rescell{1870.8}{-16.2\%}{goodgreen} & \rescell{93.99}{+0.6\%}{goodgreen} &
\rescell{1224.4}{-15.8\%}{goodgreen} &
\rescell{83.60}{+0.6\%}{goodgreen} \\

Pdrop-VP &
\rescell{586.3}{-13.5\%}{goodgreen} & \rescell{72.02}{-1.1\%}{badred} &
\rescell{2081.7}{-6.7\%}{goodgreen} & \rescell{93.98}{+0.6\%}{goodgreen} &
\rescell{1334.0}{-8.3\%}{goodgreen} &
\rescell{83.00}{-0.1\%}{badred} \\

Ll-VP &
\rescell{545.2}{-19.6\%}{goodgreen} & \rescell{73.20}{+0.5\%}{goodgreen} &
\rescell{2045.1}{-8.3\%}{goodgreen} & \rescell{93.42}{-0.0\%}{badred} &
\rescell{1295.2}{-11.0\%}{goodgreen} &
\rescell{83.31}{+0.2\%}{goodgreen} \\
\specialrule{0.6pt}{0pt}{0pt}

\rowcolor{sectiongray}
\multicolumn{7}{c}{\textbf{Retain $\le$ 5\% Tokens}} \\
\specialrule{0.6pt}{0pt}{0pt}

FastV-VP  &
\rescell{538.0}{-20.6\%}{goodgreen} & \rescell{72.54}{-0.4\%}{badred} &
\rescell{1800.1}{-19.3\%}{goodgreen} & \rescell{95.09}{+1.8\%}{goodgreen} &
\rescell{1169.0}{-19.6\%}{goodgreen} &
\rescell{83.82}{+0.8\%}{goodgreen} \\

Pdrop-VP &
\rescell{471.2}{-30.5\%}{goodgreen} & \rescell{71.63}{-1.6\%}{badred} &
\rescell{1837.1}{-17.7\%}{goodgreen} & \rescell{91.72}{-1.8\%}{badred} &
\rescell{1154.2}{-20.6\%}{goodgreen} &
\rescell{81.68}{-1.7\%}{badred} \\

Ll-VP&
\rescell{571.5}{-15.7\%}{goodgreen} & \rescell{73.85}{+1.4\%}{goodgreen} &
\rescell{1979.6}{-11.3\%}{goodgreen} & \rescell{93.43}{-0.0\%}{badred} &
\rescell{1275.6}{-12.3\%}{goodgreen} &
\rescell{83.64}{+0.6\%}{goodgreen} \\

\bottomrule[1.5pt]
\end{tabular}
}
\vspace{-0.2in}
\end{table}

\vspace{-0.1in}
\subsection{Efficiency Analysis}
\vspace{-0.05in}

\begin{figure}[t] 
    \centering
    \centerline{\includegraphics[width=0.96\linewidth]
    {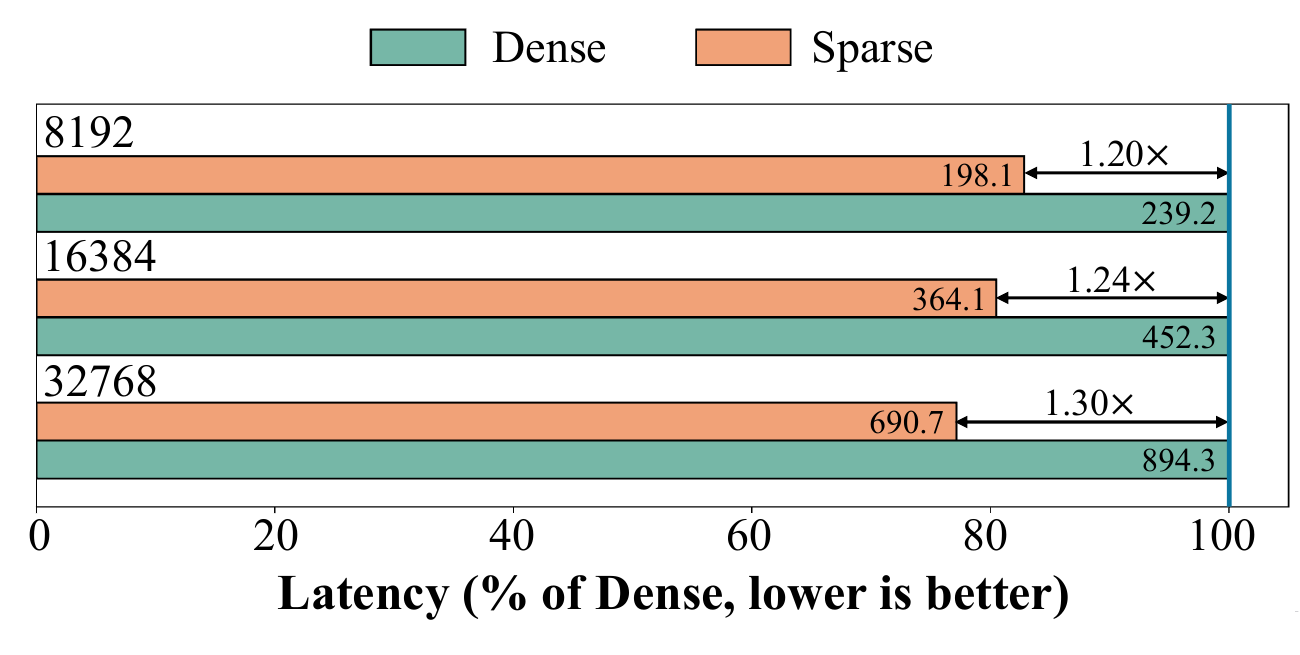}}
    % {figure/fig_5_7.pdf}}
    \vspace{-0.12in}
    \caption{\textbf{End-to-end latency comparison of dense and sparse attention under different context lengths.}
    We set the batch size to 8 and the generation length to 1k tokens. Numbers on the bars indicate the actual latency (s).}
    \label{fig:efficiency} 
    \vspace{-0.27in}
\end{figure}

Beyond reducing the generated output length, VisionPulse improves inference efficiency through maintaining dynamic visual sparsity during decoding.
Specifically, we compare the original dense attention with our dynamic visual sparse attention, and report the end-to-end latency under different context lengths.
As shown in \cref{fig:efficiency}, VisionPulse consistently achieves lower latency yielding up to \textbf{1.30$\times$} speedup over dense baseline.
Overall, by explicitly aligning the visual budget with step-wise visual dependency, VisionPulse mitigates the coupled bottleneck in multimodal reasoning and delivers stable, practical end-to-end acceleration.

\vspace{-0.1in}
\section{Conclusion}
\vspace{-0.05in}

In this paper, we revisit visual token pruning and show that visual dependency during reasoning is inherently step-dependent. 
We further identify a coupled bottleneck where redundant visual context could steer LMM toward query-irrelevant regions, yielding longer and less reliable reasoning.
To address these issues, we propose VisionPulse, a step-wise visual token pruning paradigm that estimates visual budget via a lightweight visual attention mass, and retains only critical visual tokens at each step.
Experiments show that VisionPulse retains only 5\% visual tokens per step with shortening reasoning traces, while keeping accuracy almost unchanged.
These results establish VisionPulse as an extensible framework for test-time visual compression and for studying its interplay with multimodal CoT generation.

\section*{Impact Statement}

This paper presents work whose goal is to advance the field of Machine Learning by enabling improved analysis and acceleration of complex reasoning in multimodal large language models. There are many potential societal consequences of our work, none which we feel must be specifically highlighted here.

\section*{Acknowledgements}

This work is partially supported by National Natural Science Foundation of China (62376274, 62437002). Zhiwu Lu is the corresponding author.

\bibliography{example_paper}
\bibliographystyle{icml2026}

%%%%%%%%%%%%%%%%%%%%%%%%%%%%%%%%%%%%%%%%%%%%%%%%%%%%%%%%%%%%%%%%%%%%%%%%%%%%%%%
%%%%%%%%%%%%%%%%%%%%%%%%%%%%%%%%%%%%%%%%%%%%%%%%%%%%%%%%%%%%%%%%%%%%%%%%%%%%%%%
% APPENDIX
%%%%%%%%%%%%%%%%%%%%%%%%%%%%%%%%%%%%%%%%%%%%%%%%%%%%%%%%%%%%%%%%%%%%%%%%%%%%%%%
%%%%%%%%%%%%%%%%%%%%%%%%%%%%%%%%%%%%%%%%%%%%%%%%%%%%%%%%%%%%%%%%%%%%%%%%%%%%%%%
\newpage
\appendix
\onecolumn
\section{Additional Analysis}

\subsection{Visual Attention Mass vs. Activated Token Coverage}

\begin{figure*}[h]
  % \vskip 0.2in
  \begin{center}
    \centerline{\includegraphics[width=\linewidth]{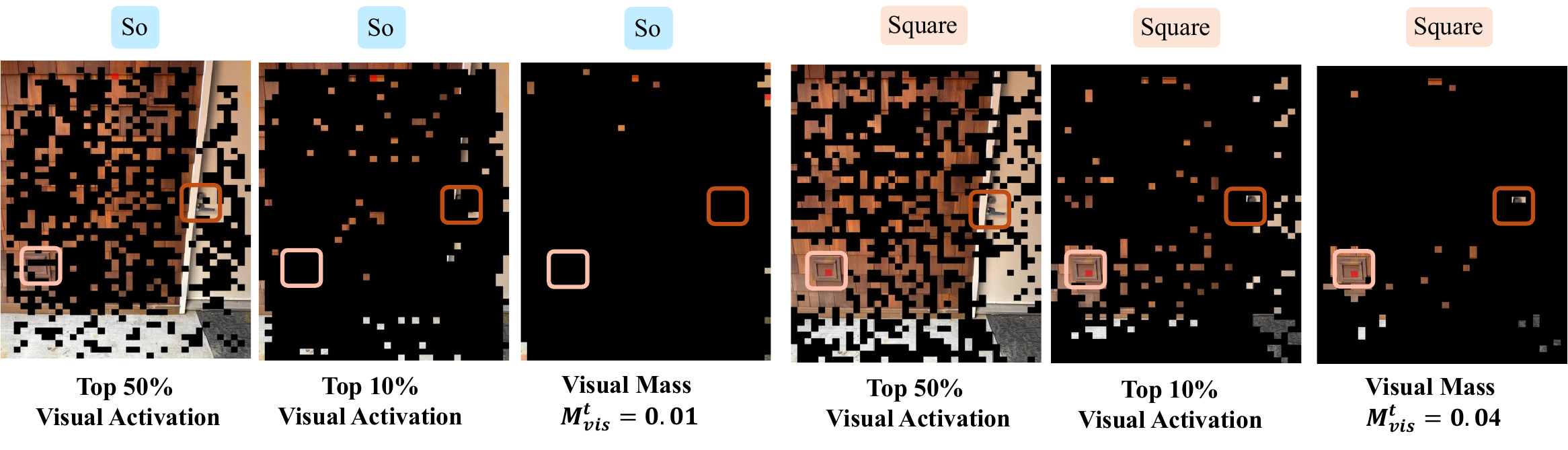}}
    \caption{\textbf{Visual attention mass vs.\ activated token coverage.} We visualize step-wise visual activation at representative decoding steps. For each step, we show the regions covered by the top-50\% and top-10\% activated visual tokens, together with the corresponding visual attention mass $M_{\mathrm{vis}}^{t}$. The examples illustrate that larger $M_{\mathrm{vis}}^{t}$ is accompanied by broader token activation, while low-mass steps exhibit sparse, localized activation.}

\label{fig:ss}
  \end{center}
  \vskip -0.25in
\end{figure*}

\cref{fig:ss} extends \cref{figure_observation} with step-wise visualizations that link the visual attention mass $M_{\mathrm{vis}}^{t}$ to the spatial coverage of activated visual tokens.
When the reasoning step is primarily language-driven (e.g., \emph{so}), $M_{\mathrm{vis}}^{t}$ is low and activation is sparse, indicating that only a small fraction of visual tokens is needed. 
In contrast, during visually grounded steps (e.g., \emph{square}), attention spreads over a broader set of tokens and higher-percentile activations cover substantially larger regions, accompanied by a clear increase in $M_{\mathrm{vis}}^{t}$.
These examples qualitatively support our finding that $M_{\mathrm{vis}}^{t}$ correlates with the number of effectively activated tokens, motivating mass-guided dynamic budgeting.

\subsection{Effects of different Budgeting strategies}

\begin{figure}[h]
  \centering
  \begin{subfigure}[t]{0.49\linewidth}
    \centering
    \includegraphics[width=\linewidth]{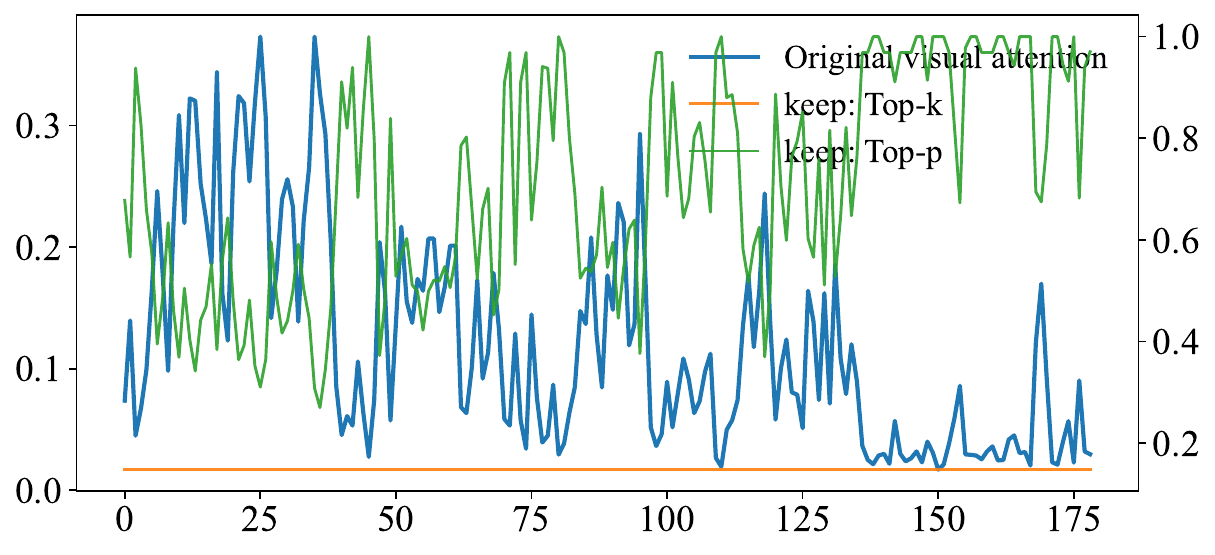}
    \caption{Top-$k$/Top-$p$ selection.}
    \label{fig:topkp_a}
  \end{subfigure}\hfill
  \begin{subfigure}[t]{0.49\linewidth}
    \centering
    \includegraphics[width=\linewidth]{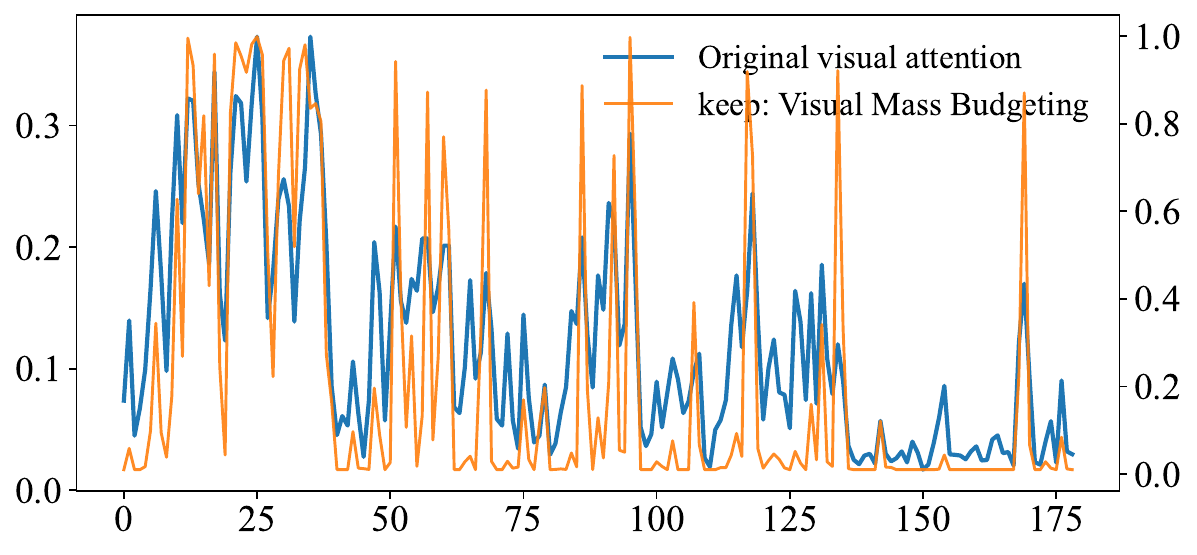}
    \caption{Visual-Mass guided selection.}
    \label{fig:topkp_b}
  \end{subfigure}
    \caption{\textbf{Budget and retention-ratio dynamics.}
    We compare the step-wise retention behavior under different budgeting strategies. 
    The blue curve shows the original visual attention mass, while the colored curves indicate the retained token ratio determined by (a) fixed Top-$K$/Top-$p$ selection and (b) Visual-Mass budgeting. 
    Visual-Mass budgeting tracks attention fluctuations more closely, allocating higher budgets at visually grounded steps and lower budgets at language-dominant steps.}
  \label{fig:topkp}
  % \vskip -0.4in
\end{figure}

We study how different budgeting strategies affect step-wise retention under a controlled setting, using Top-$p$ with $p=0.10$, Top-$k$ with $k=256$, and temperature scaling $\tau=0.4$.
As shown in~\cref{fig:topkp}, Top-$k$ and Top-$p$ strategies are poorly matched to the step-wise variation of visual dependency.
Top-$k$ allocates a constant budget across steps, and thus cannot increase retention when the model requires broader visual evidence, nor reduce retention when the step is largely language-driven.
Interestingly, Top-$p$ exhibit a trend opposite to the visual-attention variation: since softmax produces dense distributions with an inherent long tail~\cite{ren2020balanced,martins2016softmax}, even low-visual-dependency steps can accumulate non-trivial probability mass over many tokens, causing Top-$p$ to retain an unnecessarily large set and inflate the budget.
In contrast, visual mass-guided budgeting directly tracks step-wise visual demand and yields a more stable retention behavior aligned with the visual reliance.

\subsection{Complementarity with Prefill Pruning}
\label{app:persistent_inactivity}

\paragraph{Complementarity with prefill-stage token reduction.}

VisionPulse performs step-wise dynamic pruning during decoding rather than at the prefill stage, it naturally raises the question of whether it can be effectively combined with existing pruning methods applied during prefill.
Since prefill-stage pruning removes globally redundant visual tokens before decoding, it may offer additional acceleration when coupled with our decoding-time dynamic visual sparsification. 

\begin{figure}[t]
\centering
\begin{minipage}[t]{0.44\linewidth}
\vspace{-3.5cm}
\centering
\captionof{table}{Complementarity between prefill-stage token reduction and VisionPulse. Results are averaged over three benchmarks. FastV is applied with a prefill compression ratio of $0.5$, and VisionPulse further performs step-wise token sparsification during decoding.}
\vspace{0.1in}
\label{tab:fastv_vp_complementary}
\renewcommand{\arraystretch}{1.2}
\setlength{\tabcolsep}{8pt}
\scalebox{0.88}{
\begin{tabular}{lcc}
\toprule
\textbf{Method} & \textbf{Avg $L$} $\downarrow$ & \textbf{Avg $A$} $\uparrow$ \\
\midrule
FastV & 2094.7 & 69.4 \\
\textbf{FastV + VisionPulse ($\leq 10\%$)} & 1967.4 & 70.4 \\
\textbf{FastV + VisionPulse ($\leq 5\%$)} & 1957.0 & 69.8 \\
\bottomrule
\end{tabular}
}
\vspace{0.2in}
\end{minipage}
\hfill
\begin{minipage}[t]{0.53\linewidth}
\centering
\includegraphics[width=0.96\linewidth]{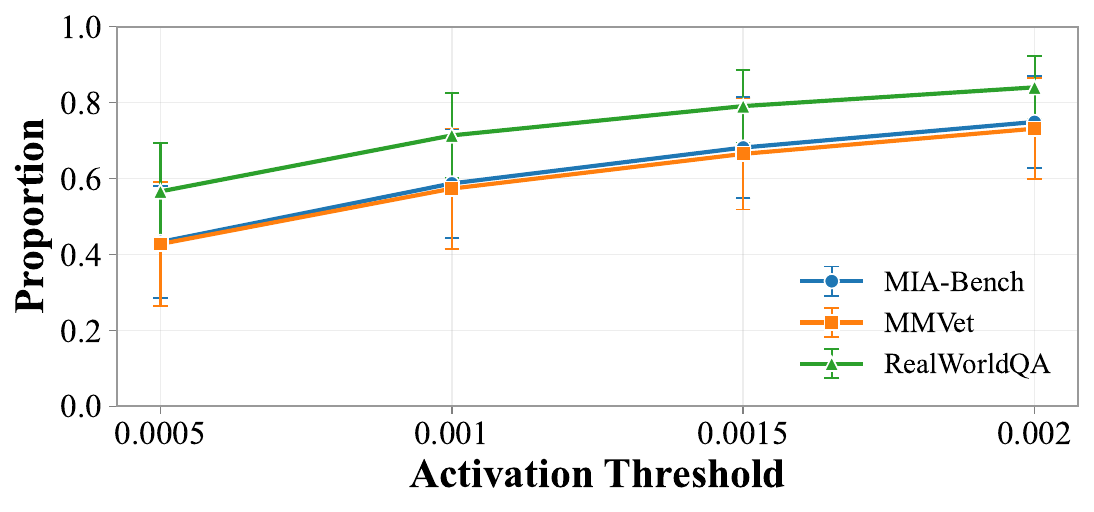}
\vspace{-0.1in}
\caption
{Proportion of never-activated visual tokens under different activation thresholds. Across MIA-Bench, MMVet, and RealWorldQA, many visual tokens remain inactive during decoding, indicating the existence of persistent visual redundancy during multimodal reasoning.}
\label{fig:absolute_inactivity}
\end{minipage}
\end{figure}

To examine this complementarity, we integrate FastV with VisionPulse and evaluate the combined method on MMVet, RealWorldQA, and MIA-Bench. Specifically, FastV is first applied during the prefill stage with a compression ratio of $0.5$, after which VisionPulse performs step-wise visual token sparsification over the remaining visual tokens during decoding. The averaged results across the three datasets are reported in Table~\ref{tab:fastv_vp_complementary}. FastV+VisionPulse consistently outperforms FastV alone, indicating that VisionPulse complements prefill-stage token reduction. This suggests that VisionPulse is orthogonal to static visual token pruning methods such as FastV and can be combined with them for additional efficiency improvements.

% \begin{table}[t]
% \centering
% \caption{Complementarity between prefill-stage token reduction and VisionPulse. Results are averaged over three benchmarks. FastV is applied with a prefill compression ratio of $0.5$, and VisionPulse further performs step-wise token sparsification during decoding.}
% \label{tab:fastv_vp_complementary}
% \renewcommand{\arraystretch}{1.2}
% \setlength{\tabcolsep}{12pt}
% \scalebox{0.88}{
% \begin{tabular}{lcc}
% \toprule
% \textbf{Method} & \textbf{Avg $L$} $\downarrow$ & \textbf{\textbf{Avg $A$} $\uparrow$} \\
% \midrule
% FastV & 2094.7 & 69.4 \\
% \textbf{FastV + VisionPulse ($\leq 10\%$)} & 1967.4 & 70.4 \\
% \textbf{FastV + VisionPulse ($\leq 5\%$)} & 1957.0 & 69.8 \\
% \bottomrule
% \end{tabular}
% }
% % \vspace{-0.1in}
% \end{table}
% \begin{figure}[t]
%     \centering
%     \includegraphics[width=0.40\linewidth]{figure/absolute_inactivity_multi_dataset.pdf}
%     \caption{Proportion of never-activated visual tokens under different activation thresholds. Across ChartQA, MIA-Bench, MMVet, and RealWorldQA, a large fraction of visual tokens remain inactive throughout decoding, indicating the existence of persistent visual redundancy during multimodal reasoning.}
%     \label{fig:absolute_inactivity}
%     % \vspace{-0.2in}
% \end{figure}

\paragraph{Persistent visual redundancy during reasoning.}

Beyond verifying the complementarity between VisionPulse and prefill-stage pruning, we further examine the temporal behavior of visual token usage during decoding. 
Specifically, we explore whether some visual tokens remain consistently inactive throughout the reasoning process. 
For each visual token, we measure its attention activation across all decoding steps and regard it as persistently inactive if its activation remains below a threshold $\delta$ at every step. 
As shown in Figure~\ref{fig:absolute_inactivity}, across different datasets and threshold choices, a substantial fraction of visual tokens are rarely or never activated during reasoning.
This observation suggests that multimodal reasoning contains a persistent inactive visual subset, which in principle can be removed without affecting the reasoning process.
However, existing one-shot prefill-stage pruning methods are not specifically designed to exploit this temporal structure.
These methods make the pruning decision before decoding begins, relying primarily on an initial static estimate of token importance to determine the retained visual subset.
As a result, the retained subset may still include visual tokens that remain inactive in later reasoning steps, while some discarded tokens may become relevant as the decoding process evolves.

\section{Implementation Details}
\label{sec:appendix_impl}

\subsection{Experimental Setup}
\label{sec:experimental_setup}
We mainly evaluate VisionPulse on Qwen3-VL-Thinking-4B under aggressive visual-token retention. 
Unless otherwise stated, all results are obtained with decoding temperature $0.7$ and a maximum generation length of $16{,}384$ tokens.
In \cref{eq6}, we set the temperature scaling parameter $\tau$ to $0.4$ for the $10\%$ retention setting and to $0.1$ for the $5\%$ retention setting.
For each benchmark, we follow the standard evaluation protocol and report accuracy (\%) together with the average generation length $L$.
All inference experiments are conducted on 8× NVIDIA A800 80GB GPUs.

\subsection{Datasets and Evaluation Protocols}
\label{sec:appendix_datasets}
We evaluate on seven multimodal reasoning benchmarks: CharXiv RQ~\cite{charxiv}, InfoVQA~\cite{infovqa}, ChartQA~\cite{chartqa}, MMStar~\cite{mmstar}, RealWorldQA~\cite{realworldqa}, MMVet~\cite{mmvet}, and MIA-Bench~\cite{miabench}.
All evaluations are conducted with VlMEvalKit~\cite{duan2024vlmevalkit} using the default prompts and scoring rules.
For LLM-judged benchmarks, we use GPT-4o for MIA-Bench and GPT-4o-mini for MMVet, CharXiv RQ, and MMStar.

\textbf{CharXiv RQ.}~
CharXiv~\cite{charxiv} is a realistic chart-understanding benchmark curated from scientific papers, comprising 2,323 natural charts with associated questions.
It emphasizes fine-grained visual grounding and domain-specific comprehension, covering both descriptive chart reading and reasoning-intensive queries.
We use the reasoning validation subset as CharXiv RQ to evaluate the performance of our method, follow the default CharXiv evaluation in VlMEvalKit.

\textbf{InfoVQA.}~
InfoVQA (InfographicVQA)~\cite{infovqa} focuses on question answering over infographics that required methods to jointly reason over the document layout, textual content, graphical elements, and data visualizations.
It contains 5,485 infographic images and 30,035 question--answer pairs, with diverse sources and strong layout/OCR dependence.
InfoVQA is widely used to evaluate visual-dominant multimodal document understanding that requires integrating reading, layout cues, and visual symbols.
We follow the default InfoVQA evaluation in VlMEvalKit.

\textbf{ChartQA.}~
ChartQA~\cite{chartqa} tests question answering about charts with both visual and logical reasoning, often requiring numerical comparison and aggregation grounded in chart elements.
It includes 9.6K human-authored questions and 23.1K machine-generated questions focusing on logical and visual reasoning questions, which involve complex reasoning and visual references to charts.
We follow the default ChartQA evaluation in VlMEvalKit.

\begin{figure*}[t]
  % \vskip 0.2in
    \centering
    \centerline{\includegraphics[width=\linewidth]{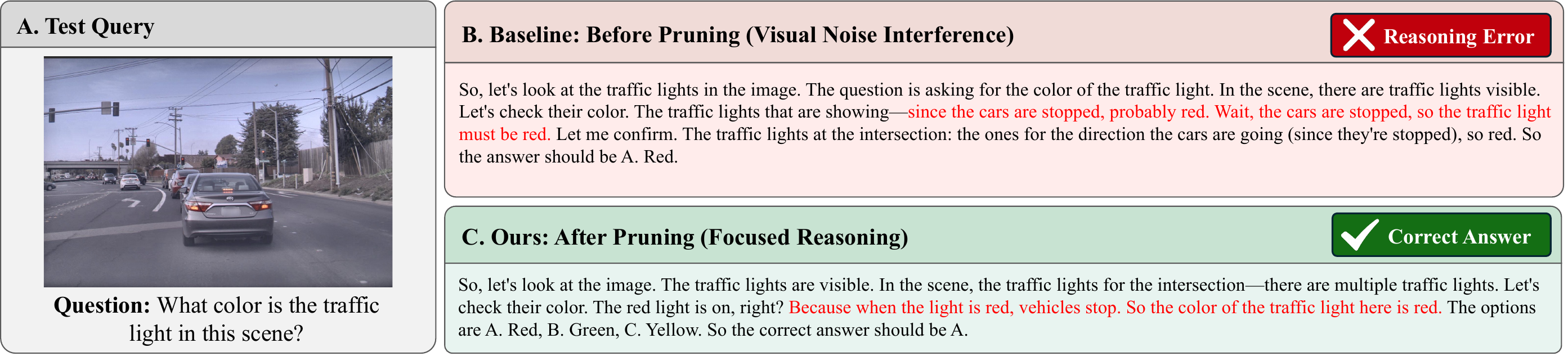}}
    \caption{\textbf{Qualitative example of visual-noise interference on unnecessary reasoning.}
    Given the same query, the full-context baseline is distracted by query-irrelevant visual cues (e.g., stopped cars) and produces an incorrect reasoning trace, whereas VisionPulse (C) prunes redundant visual tokens during decoding, focuses attention on the traffic light, and yields the correct answer.}
    \label{fig:example1}
    \vskip -0.2in
\end{figure*}

\begin{figure*}[t]
  % \vskip 0.2in
    \centering
    \centerline{\includegraphics[width=\linewidth]{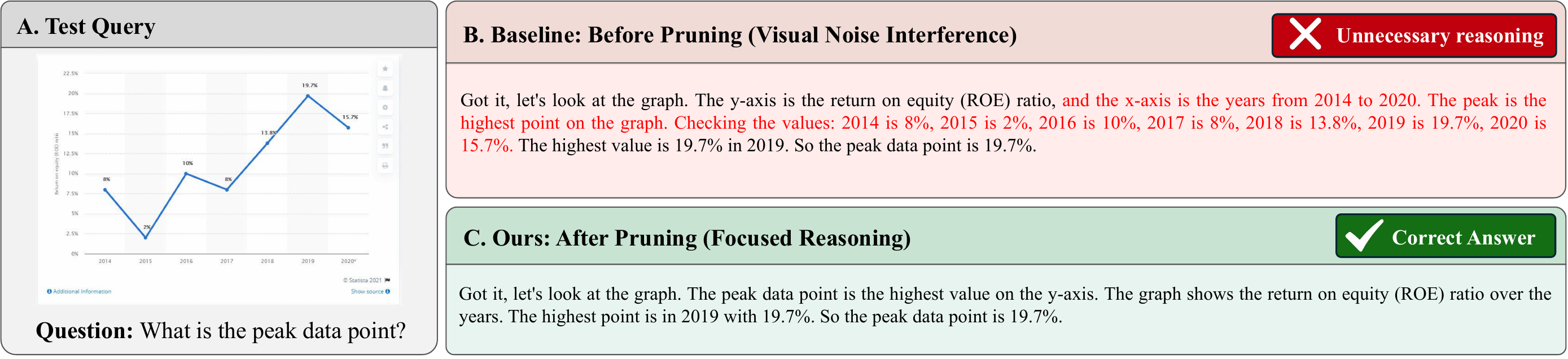}}
    \caption{\textbf{Qualitative example of visual-noise interference on unnecessary reasoning.} On chart-based queries, the full-context baseline (B) tends to produce verbose, query-irrelevant descriptions by enumerating many values in the plot, resulting in unnecessary reasoning and longer generation. In contrast, VisionPulse (C) prunes redundant visual context during decoding, keeps attention on the peak point required by the question, and answers correctly with a shorter, more focused reasoning trace.}
    \label{fig:example3} 
    \vskip -0.1in
\end{figure*}

\begin{figure*}[t]
    \centering
    \centerline{\includegraphics[width=\linewidth]{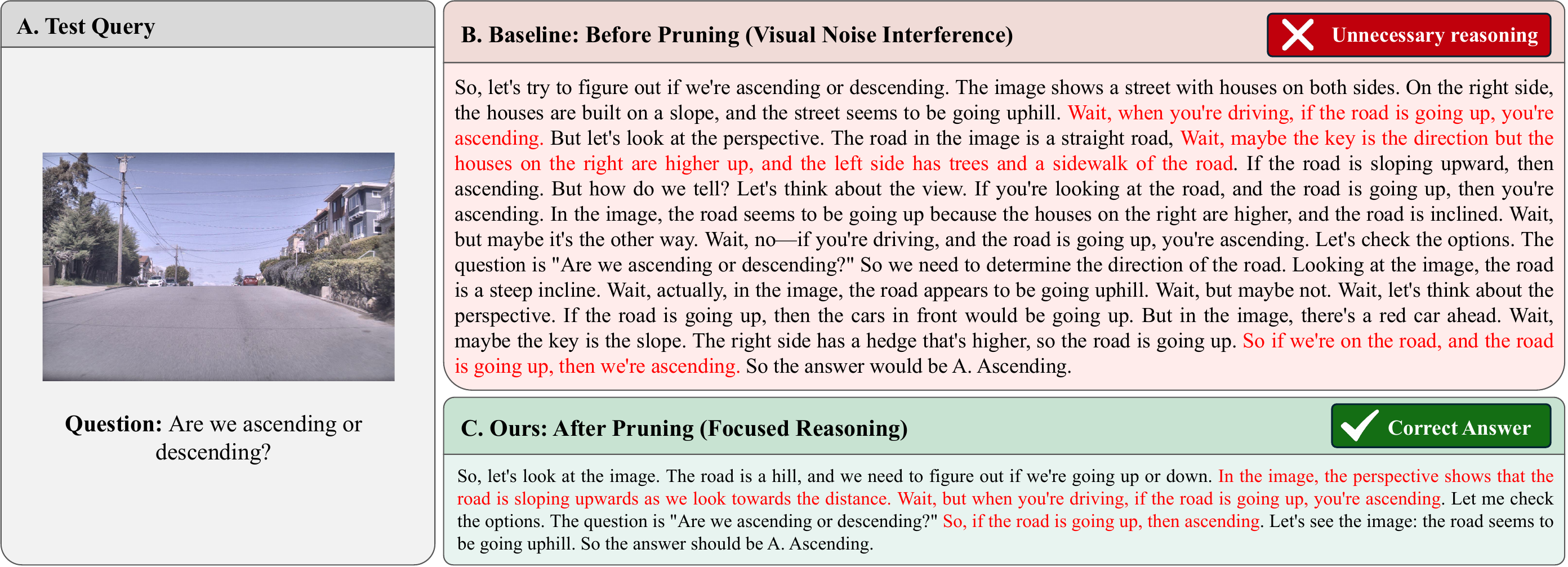}}
    \caption{\textbf{Qualitative example of redundant visual reasoning.} In this example, the LMM identifies the correct reasoning direction early on, but is later distracted by query-irrelevant visual cues (e.g., buildings and trees), which triggers repetitive reasoning. As a result, it follows a prolonged trajectory before eventually returning to the same conclusion implied by its initial reasoning.}
    \label{fig:example2} 
    \vskip -0.2in
\end{figure*}

\textbf{MMStar.}~
MMStar~\cite{mmstar} is a vision-indispensable benchmark designed to reduce shortcut solving and ensure that visual evidence is necessary for answering.
It comprises 1,500 human-curated challenge samples organized into 6 core capabilities and 18 fine-grained axes for diagnostic evaluation.
Strict human review is involved to ensure each selected sample exhibits visual dependency, minimal data leakage, and requires advanced multi-modal capabilities for the solution.
We follow the default MMStar evaluation in VlMEvalKit.

\textbf{RealWorldQA.}~
RealWorldQA~\cite{realworldqa} evaluates real-world visual understanding with an emphasis on basic spatial and physical reasoning.
The initial release contains over 700 images, each paired with a question and an easily verifiable answer, including anonymized vehicle-captured scenes.
It is commonly used to test grounding in everyday real-world contexts rather than memorized or purely language-driven behavior.
We follow the default RealWorldQA evaluation in VlMEvalKit.

\textbf{MMVet.}~
MMVet~\cite{mmvet} is an open-ended benchmark designed to evaluate integrated vision-language skills on complicated multimodal tasks, such as recognition, OCR, spatial reasoning, knowledge, and arithmetic
It contains 200 images and 218 questions, each annotated with the capabilities required for solving.
We follow the default MMVet evaluation in VlMEvalKit.

\textbf{MIA-Bench.}~
MIA-Bench~\cite{miabench} focuses on complex multimodal instruction following, emphasizing strict adherence to layered, constraint-heavy prompts in open-ended generation.
It comprises 400 image--prompt pairs crafted to elicit instruction violations (e.g., formatting constraints, multi-part requirements, and compositional conditions).
MIA-Bench is commonly used to evaluate controllability and instruction fidelity of MLLMs under complex prompts.
We follow the default MIA-Bench evaluation in VlMEvalKit.

\section{More Qualitative Analysis for The Coupled Bottleneck in LMMs}
\label{appendix:examples}
In this section, we provide additional qualitative examples (\cref{fig:example1,fig:example2,fig:example3}) to analyze how query-irrelevant visual evidence induces unnecessary reasoning and contributes to the coupled bottleneck in LMMs.

\section{More Qualitative Analysis of Different Pruning Paradigms}

\begin{figure*}[t]
    \centering
    \centerline{\includegraphics[width=\linewidth]{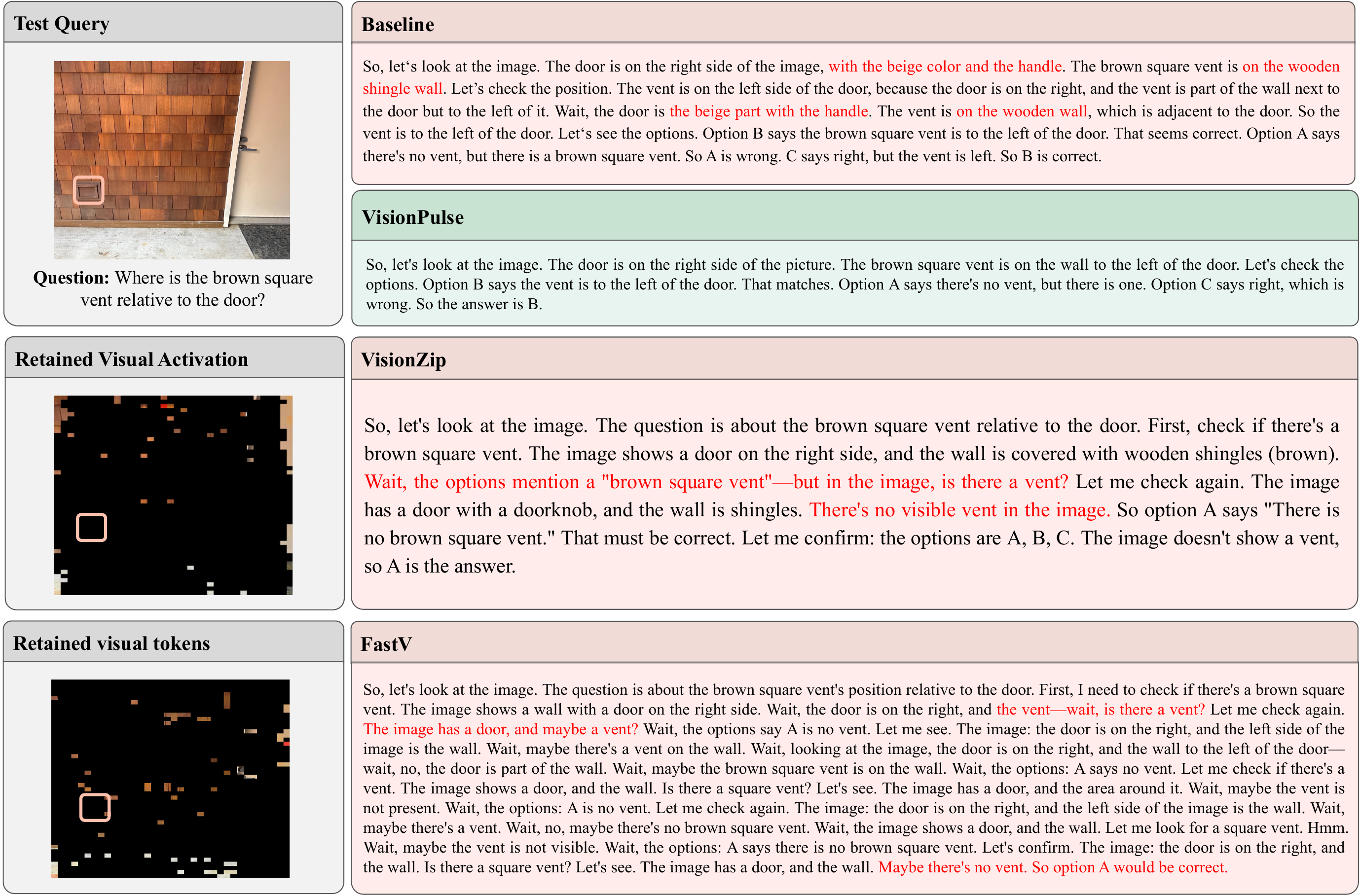}}
    \caption{\textbf{Visualization comparison of the reasoning trajectories of different visual token pruning methods.} We further visualize the retained visual tokens of VisionZip (dominant tokens) and FastV under 5\% visual token retention.}
    \label{fig:diff_pruning} 
    \vskip -0.2in
\end{figure*}

To provide a more intuitive qualitative analysis of different pruning paradigms, we visualize both the reasoning trajectories and the retained visual tokens under 5\% visual token retention. 
As shown in~\cref{fig:diff_pruning}, static pruning methods struggle to preserve task-relevant visual information, making them inadequate for multimodal reasoning and often leading to incorrect predictions. 
In contrast, VisionPulse dynamically retains the visual tokens required at each reasoning step, thereby preserving critical information and enabling correct reasoning.
This observation is also consistent with our discussion that, during the prefill stage, the model tends to prioritize generating the first token rather than acquiring all the visual evidence required for subsequent reasoning.

\end{document}